\pdfoutput=1
\documentclass[11pt]{article}
\usepackage[]{acl}
\usepackage{times}
\usepackage{latexsym}
\usepackage[T1]{fontenc}
\usepackage[utf8]{inputenc}
\usepackage{microtype}
\usepackage{inconsolata}
\usepackage{booktabs}
\usepackage{graphicx}
\usepackage{makecell}
\usepackage{multirow}
\usepackage{colortbl}
\usepackage{float}
\usepackage{xcolor}
\usepackage{amsmath}
\usepackage{amssymb}
\usepackage{textcomp}
\usepackage{pifont}
\usepackage{cancel}
\usepackage{enumitem}
\usepackage{courier}

\definecolor{blue}{RGB}{50,80,130}
\definecolor{red}{RGB}{120,30,35}
\definecolor{green}{RGB}{34,85,50}
\definecolor{gray}{RGB}{90,100,110}
\definecolor{cyan}{RGB}{70,120,140}
\definecolor{purple}{RGB}{150,110,140}
\definecolor{softred}{RGB}{200,90,95}
\definecolor{softgreen}{RGB}{85,150,100}

\title{
  \textcolor{blue}{\textit{What Are They Talking About?}} \\
  A Benchmark of Knowledge-Grounded Discussion Summarization
}

\author{Weixiao Zhou\textsuperscript{\textnormal{$\alpha$}} \quad
		Junnan Zhu\textsuperscript{\textnormal{$\beta$}} \quad
		Gengyao Li\textsuperscript{\textnormal{$\beta\gamma$}} \\
		\textbf{Xianfu Cheng}\textsuperscript{\textnormal{$\alpha$}} \quad
		\textbf{Xinnian Liang}\textsuperscript{\textnormal{$\delta$}} \quad
		\textbf{Feifei Zhai}\textsuperscript{\textnormal{$\beta\epsilon$}} \quad
		\textbf{Zhoujun Li}\textsuperscript{\textnormal{$\alpha$}}\thanks{Corresponding Author} \\
        \textsuperscript{\textnormal{$\alpha$}}State Key Laboratory of Complex \& Critical Software Environment, Beihang University \\
        \textsuperscript{\textnormal{$\beta$}}State Key Laboratory of Multimodal Artificial Intelligence Systems, Institute of Automation, CAS \\
        \textsuperscript{\textnormal{$\gamma$}}University of Chinese Academy of Sciences \quad
        \textsuperscript{\textnormal{$\delta$}}ByteDance Inc. \quad
        \textsuperscript{\textnormal{$\epsilon$}}Fanyu AI Laboratory \smallskip \\
        {\fontsize{11pt}{0pt}\selectfont \texttt{wxzhou@buaa.edu.cn} \quad \texttt{junnan.zhu@nlpr.ia.ac.cn}}}

\begin{document}
\maketitle

\begin{abstract}
Traditional dialogue summarization primarily focuses on dialogue content, assuming it comprises adequate information for a clear summary. However, this assumption often fails for discussions grounded in shared background, where participants frequently omit context and use implicit references. This results in summaries that are confusing to readers unfamiliar with the background. To address this, we introduce \textbf{K}nowledge-\textbf{G}rounded \textbf{D}iscussion \textbf{S}ummarization (\textbf{KGDS}), a novel task that produces a supplementary \textit{background summary} for context and a clear \textit{opinion summary} with clarified references. To facilitate research, we construct the first KGDS benchmark, featuring news-discussion pairs and expert-created multi-granularity gold annotations for evaluating sub-summaries. We also propose a novel hierarchical evaluation framework with fine-grained and interpretable metrics. Our extensive evaluation of 12 advanced large language models (LLMs) reveals that KGDS remains a significant challenge. The models frequently miss key facts and retain irrelevant ones in background summarization, and often fail to resolve implicit references in opinion summary integration.\footnote{Our benchmark is available at \href{https://github.com/zhouweixiao/KGDS}{\texttt{zhouweixiao/KGDS}}}
\end{abstract}

\section{Introduction}
Dialogue summarization aims to distill key topics and interactions from a dialogue into a concise summary \cite{kirstein2024cads, rennard2023abstractive, jia2023taxonomy}. Conventional paradigm relies primarily on dialogue content, whether in benchmark construction \cite{gliwa-etal-2019-samsum, chen-etal-2021-dialogsum, zhu-etal-2021-mediasum}, methodologies \cite{zhou-etal-2023-multi, tian-etal-2024-dialogue, lu2025mutual, zhu-etal-2025-factual}, or evaluation \cite{wang-etal-2022-analyzing, gao-wan-2022-dialsummeval, zhu-etal-2023-annotating, tang-etal-2023-context, tang2024tofueval, liu2024exploring, ramprasad2024analyzing}. These efforts implicitly assume that "\textit{the dialogue itself contains sufficient information to generate a clearly understandable summary for readers.}"

\begin{figure*}[t]
  \centering
  \setlength{\abovecaptionskip}{5pt}
  \includegraphics[width=0.977\linewidth]{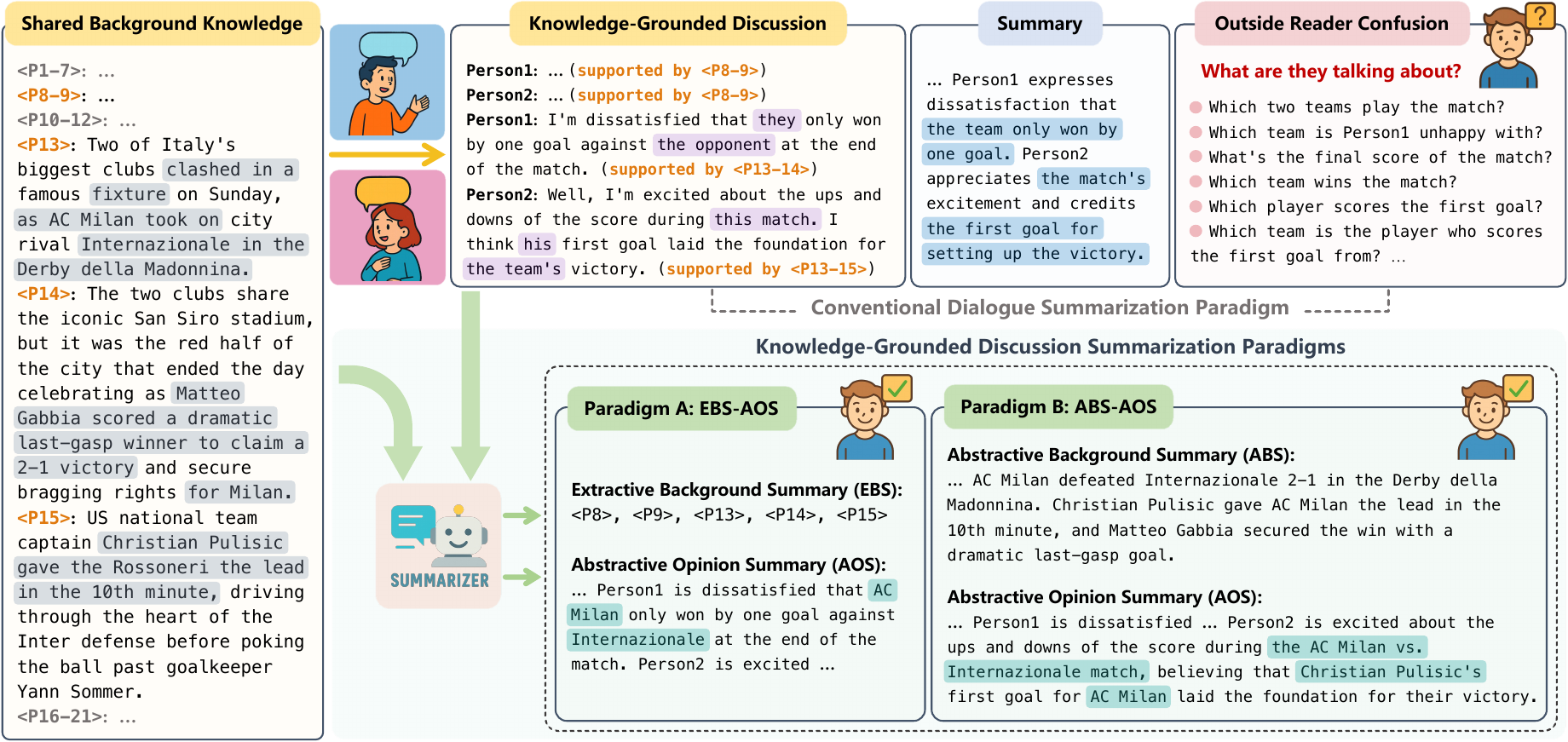}
  \caption{An overview example. \textcolor{gray}{\textbf{Gray Blocks}} in shared background knowledge denote crucial background details omitted by participants during discussion. \textcolor{purple}{\textbf{Purple Blocks}} in discussion content indicate referential pronouns or phrases. \textcolor{blue}{\textbf{Blue Blocks}} in discussion summary represent content that may cause confusion for outside readers. Compared to traditional dialogue summarization, KGDS achieves \textbf{\textit{better reader preference}} by providing a supplementary background summary and a clear opinion summary, in which \textcolor{cyan}{\textbf{Cyan Blocks}} highlight clarified implicit references.}
  \label{fig:introduction_overall}
  \vspace{-10pt}
\end{figure*}

However, we find this assumption has fundamental limitations and often fails, particularly when \textit{participants discuss shared background knowledge they are already familiar with}. Such discussions exhibit two main traits: (1) Information Omission and Implicit Reference: Participants naturally skip mutually known details and frequently use pronouns or phrases to refer to entities or facts within the contextual background. (2) Personal Opinion: Unlike simple information interactions in general dialogues, these discussions focus on exchanging viewpoints, with participants expressing personal opinions from various perspectives. These characteristics make understanding the discussion heavily reliant on background knowledge. Consequently, traditional dialogue summarization paradigm inherits this dependency, leading to confusion for outside readers unfamiliar with the context, leaving them wondering: "\textbf{What are they talking about?}". We present an illustrative example in Figure~\ref{fig:introduction_overall}.

To address this problem, we introduce KGDS, a novel task designed to combine shared background knowledge with discussion content to create reader-centered summaries. We argue that a successful KGDS summary must achieve two complementary objectives: (1) to bridge readers' knowledge gaps by providing the necessary background information that supports the discussion; and (2) to present readers with clear participant opinions by clarifying the implicit references within them. Thus, we formalize these requirements by modeling the task output as a \textbf{Background Summary} and an \textbf{Opinion Summary}. The background summary, which retrieves or condenses relevant background information, can be \textbf{either Extractive or Abstractive}, while the opinion summary integrates clarified opinions and is \textbf{inherently Abstractive}. In Figure~\ref{fig:introduction_overall}, we provide a visual example of our task paradigms.

To advance research in this field, we construct the first KGDS benchmark, situated in a common and realistic scenario of news discussions. It consists of event-rich news articles, each paired with a human-authored discussion. To enable robust assessment, we develop multi-granularity gold evaluation components, which are annotated by experts under strict consistency controls. For the extractive background summary, we annotate coarse-grained \textit{supporting} and \textit{nonsupporting paragraphs}. For the abstractive background summary, we create two sets of finest-grained atomic facts \cite{min-etal-2023-factscore}, including \textit{key supporting} and \textit{nonsupporting facts}. For the opinion summary, we introduce \textit{clear atomic opinions} as basic evaluation units that are minimized and have implicit references clarified.

Furthermore, we propose a novel hierarchical framework for evaluation. For each summarization paradigm, we first evaluate its sub-summaries separately and then aggregate their scores to derive a paradigm-level score. At the sub-summary level, our framework assesses multiple dimensions. For the two types of background summaries, we evaluate their \textbf{coverage}, \textbf{focus}, and \textbf{overall quality} by comparing them against annotated paragraphs or atomic facts, using paragraph index matching or LLM-based fact verification \cite{wei2024longformfactualitylargelanguage}. For the opinion summary, we assess \textbf{overall quality} by checking its coverage of annotated atomic opinions and identify fine-grained \textbf{integration errors} by classifying failure types for uncovered opinions. We also conduct human evaluation and show a strong correlation with our automatic metrics.

We evaluate 12 advanced LLMs on our benchmark under the \textit{structured-prompt} \cite{li2023structuredchainofthoughtpromptingcode} and \textit{self-reflection} \cite{NEURIPS2023_1b44b878} settings. Our comprehensive results demonstrate that KGDS remains a substantial challenge, with even the top-performing models achieving an average score below 69\%. We also identify several key weaknesses: LLMs struggle with the coverage-focus trade-off in background summary retrieval, frequently miss key facts during background summary generation, and fail to resolve implicit references in opinion summary integration. Furthermore, the limited effectiveness of self-reflection indicates that current LLMs lack sufficient self-correction abilities for this task. These findings highlight key bottlenecks and provide concrete directions for future advancements in coarse-grained retrieval, fine-grained generation, and knowledge integration.

\section{Task Formulation}
\label{sec:task_formulation}
Let $K$ denote the shared background knowledge among participants and $D$ represent the discussion grounded in $K$. The objective of KGDS is to provide a supplementary background summary and a clear opinion summary by integrating $K$ and $D$. We define two summarization paradigms based on the type of background summary.

\paragraph{EBS-AOS Paradigm.} The output comprises an extractive background summary (EBS) and an abstractive opinion summary (AOS):
\begin{equation}
B_e, O \leftarrow f(K, D), \; B_e \subseteq K
\label{eq:ebs-aos}
\end{equation}
Here, $f$ is the summarizer. $B_e$ is the EBS, defined as extractive background supporting chunks for $D$ from $K$. Chunks are text sequences of a predefined granularity (e.g., sentences, paragraphs, etc.). $O$ is the AOS, defined as clear personal opinions of the participants with clarified implicit references.

\paragraph{ABS-AOS Paradigm.} The output contains an abstractive background summary (ABS) and an abstractive opinion summary (AOS):
\begin{equation}
B_a, O \leftarrow f(K, D)
\end{equation}
The definitions of $f$ and $O$ follow Eq.~\eqref{eq:ebs-aos}. $B_a$, the ABS, is defined as abstractive background supporting information for $D$ from $K$.

\section{Benchmark Construction}
\subsection{Preliminary}
\paragraph{Scenario Setting.} We establish our benchmark scenario as a two-participant discussion of news content for two reasons. First, news discussions are highly prevalent in daily life, making them more representative than private scenarios such as internal meetings and medical consultations. Second, news summarization \cite{goyal2022news, zhang2024benchmarking, liu-etal-2024-benchmarking} is a well-established subfield of automatic summarization research.

\paragraph{News Collection.} We collect 100 multi-domain (i.e., business, sports, and world) event-rich news articles from Google News\footnote{\href{https://news.google.com}{\texttt{https://news.google.com}}} as shared background knowledge, with a time cutoff of Oct. 2024. We preserve the original news paragraph structure and define \textit{\textbf{paragraph-as-chunk}} as the minimum extraction granularity under the EBS-AOS paradigm.

\paragraph{Expert Annotators.} To ensure high-quality data, we recruit four PhD candidates specializing in NLP for our annotation tasks. Each pair of experts forms a collaborative group to conduct the full-process annotation, which includes discussion generation and the creation of multi-granularity evaluation components for background and opinion summaries.

\subsection{Annotation}
\paragraph{Human Discussion.} This construction follows the sequence of \textbf{read, understand, then discuss}. For each news article, we require two experts to independently read and thoroughly comprehend its content. This preparatory step aims to ensure background consistency between the participants. Afterward, they engage in a discussion to exchange viewpoints. The entire process is open-ended, meaning the discussion initiator is chosen randomly, and the discussion topics can encompass any events, facts, or detailed information within the news article.

\begin{table}[t]
    \centering
    \small
    \setlength{\abovecaptionskip}{7pt}
    \setlength{\tabcolsep}{6pt}
    \begin{tabular}{lcc}
        \toprule[1pt]
        \textbf{Data} & \textbf{Scale} & \textbf{Avg. Tokens} \\
        \midrule[0.5pt]
        \rowcolor{gray!20}
        \multicolumn{3}{l}{\textit{KGDS Task Inputs}} \\
        News-Discussion Pair & 100 & 729.5 \\
        \midrule[0.5pt]
        \rowcolor{gray!20}
        \multicolumn{3}{l}{\textit{EBS Evaluation Components}} \\
        Supporting Paragraph & 432 & 43.8 \\
        Nonsupporting Paragraph & 1,005 & 42.6 \\
        \midrule[0.5pt]
        \rowcolor{gray!20}
        \multicolumn{3}{l}{\textit{ABS Evaluation Components}} \\
        Key Supporting Atomic Fact & 1,638 & \cellcolor{softgreen!25}11.3 \\
        Nonsupporting Atomic Fact & 4,996 & \cellcolor{softgreen!25}11.4 \\
        \midrule[0.5pt]
        \rowcolor{gray!20}
        \multicolumn{3}{l}{\textit{AOS Evaluation Components}} \\
        Clear Atomic Opinion & 873 & \cellcolor{softgreen!15}19.4 \\
        Clarified Implicit References & 1,113 & \cellcolor{softgreen!35}4.1 \\
        \bottomrule[1pt]
    \end{tabular}
    \caption{Benchmark statistics. For three types of sub-summaries, we annotate evaluation components at multiple granularities, including coarse-grained paragraphs, fine-grained facts and opinions. The green boxes highlight the average lengths of fine-grained annotations.}
    \label{tab:benchmark_statistics}
    \vspace{-7pt}
\end{table}

\vspace{5pt}
\noindent{\textbf{Paragraph-Level Annotation for EBS.}} \quad \enspace An ideal EBS should include all paragraphs that support the discussion and exclude any that do not. To create gold labels for this, two experts independently classify each paragraph from the source news as either \textit{supporting} or \textit{nonsupporting}. We perform a consistency control, retaining only paragraphs where both experts agree on the label. Paragraphs with conflicting annotations are considered \textit{ambiguous} and are removed from the news source to ensure data clarity. Our statistics show that out of a total of 1,696 paragraphs, 1,437 (84.7\%) are annotated consistently. Table~\ref{tab:benchmark_statistics} provides more statistics.

\vspace{5pt}
\noindent{\textbf{Atomic-Fact-Level Annotation for ABS.}} \quad Unlike coarse-grained paragraph annotation, a model-generated ABS is condensed and requires fine-grained, key-point-focused labeling. Inspired by the \textit{minimal granularity} of atomic facts \cite{liu-etal-2023-revisiting, tang-etal-2024-minicheck}, we argue that an ideal ABS should cover all \textit{key supporting} facts while filtering out \textit{nonsupporting} ones. Motivated by this, we create these two distinct atomic fact sets.

First, we utilize GPT-4o\footnote{\texttt{gpt-4o-2024-08-06}} to decompose supporting paragraphs into candidate atomic facts based on principles of indivisibility, independence, and declarativity (see Appendix~\ref{sec:atomic_fact_decomposition_instruction}). Two experts then independently classify each fact as either key or non-key. Through a consistency check, only facts unanimously labeled as key by both experts are considered as the gold standard, resulting in 1,638 key supporting atomic facts.

\begin{table}[t]
    \centering
    \small
    \setlength{\abovecaptionskip}{5pt}
    \begin{tabular}{l}
        \toprule[1pt]
        Person1 thinks that it was wise of \textcolor{green}{\textbf{Rees-Zammit}} (\textcolor{red}{\textbf{him}}) \\
        to sign with \textcolor{green}{\textbf{the Kansas City Chiefs}} (\textcolor{red}{\textbf{this team}}). \\
        \midrule[0.5pt]
        Person2 thinks that worrying about \textcolor{green}{\textbf{Jersey's}} (\textcolor{red}{\textbf{their}}) \\
        economic diversification is an overconcern. \\
        \bottomrule[1pt]
    \end{tabular}
    \caption{Examples of clear atomic opinions. Each opinion is an indivisible fine-grained basic unit with clarified implicit references. The content in parentheses indicates the original referential pronouns and phrases.}\label{tab:clear_atomic_opinion}
    \vspace{-10pt}
\end{table}

Second, we use the same decomposition method to nonsupporting paragraphs to generate an initial set of nonsupporting facts. From this set, we then mask\footnote{The masking process is consistent with automatic fact verification \cite{tang-etal-2024-minicheck}, and the masked conflicting facts are not considered in ABS evaluation.} any \textit{conflicting} facts, defined as those that could also be inferred from supporting paragraphs. Such conflicts typically arise when identical or similar facts appear in both supporting and nonsupporting paragraphs at the atomic granularity level. For instance, two events have the same timestamp, but one is the background event while the other is not. By masking these facts, we ensure that the key supporting and nonsupporting fact sets remain non-overlapping, thereby avoiding external verification issues during evaluation. This process yields 4,996 nonsupporting atomic facts, with conflicting facts proving to be sparse at just 176 instances (3.39\%).

\vspace{6pt}
\noindent{\textbf{Atomic-Opinion-Level Annotation for AOS.}} \quad According to the task setup, an effective AOS must clearly present participant opinions by clarifying implicit references. To enable fine-grained evaluation, we introduce the \textit{clear atomic opinion} as the basic unit, which is both \textit{minimal} and has \textit{implicit references clarified}. Table~\ref{tab:clear_atomic_opinion} provides two examples. We argue that the quality of an AOS can be assessed by measuring its coverage of these atomic opinions. For annotation, we first require experts to extract the main opinions from their respective utterances in the discussion. They then identify referential pronouns and phrases within these opinions and clarify them through anaphora resolution or information supplementation to produce clear opinions. Finally, the experts decompose these clear opinions into atomic opinion units, following the principles of indivisibility and independence. This annotation process is fully expert-authored. We create 873 clear atomic opinions, among which 800 contain clarified implicit references, while 73 require no clarification. Table~\ref{tab:benchmark_statistics} provides statistics.

\begin{figure*}[t]
  \centering
  \setlength{\abovecaptionskip}{5pt}
  \includegraphics[width=0.977\linewidth]{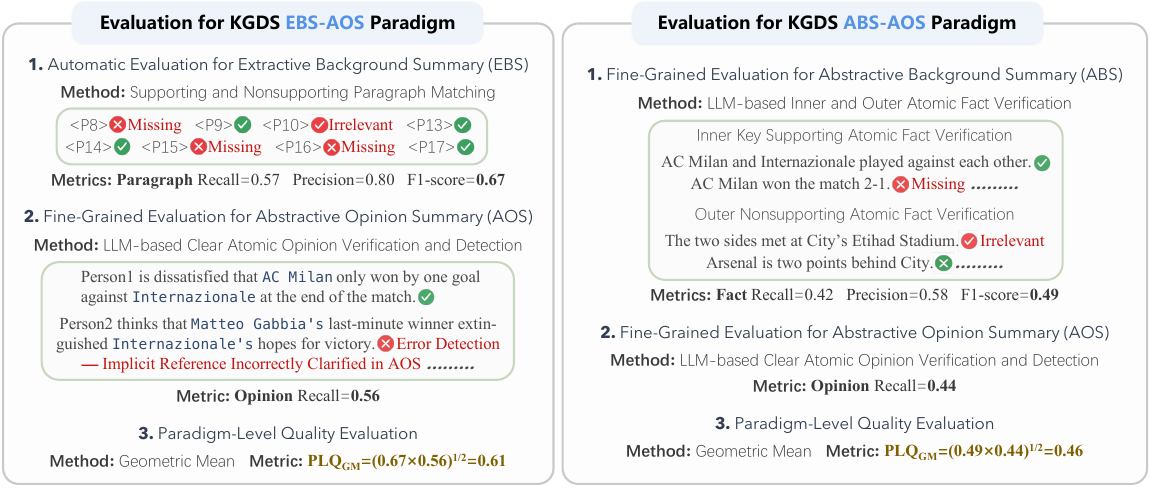}
  \caption{An overview example of our evaluation framework. It comprehensively and accurately evaluates sub-summaries and summarization paradigms through fine-grained, interpretable metrics and hierarchical aggregation.}
  \label{fig:evaluation_framework}
  \vspace{-10pt}
\end{figure*}

\section{Evaluation Framework}
Our framework is comprehensive and hierarchical. For each of the two summarization paradigms (\S\ref{sec:task_formulation}), we independently evaluate the sub-summaries and then aggregate their quality to assess overall performance at the paradigm level. The evaluation methods are interpretable, and the metrics are fine-grained. Figure~\ref{fig:evaluation_framework} shows an example. Below, we first describe the evaluation dimensions\footnote{We do not focus on \textit{fluency} and \textit{coherence} because current LLMs perform well on these dimensions \cite{song2025learningsummarizellmgeneratedfeedback}.} at the sub-summary and paradigm levels (\S\ref{sec:evaluation_dimensions}), and then introduce our automatic methods and metrics (\S\ref{sec:methods_and_metrics}).

\subsection{Dimensions}
\label{sec:evaluation_dimensions}
\vspace{2pt}
\paragraph{Background Summary.} For the two variants of background summaries, we measure the following dimensions of system responses:
\begin{itemize}[leftmargin=12pt, itemsep=0pt, topsep=2pt]
  \item \textbf{Coverage}: Whether the model-extracted or -generated summary fully covers the supporting information (i.e., paragraphs or key facts).
  \item \textbf{Focus}: Whether the summary focuses on supporting information while excluding nonsupporting information.
  \item \textbf{Overall Quality}: A holistic assessment of the summary, considering both coverage and focus.
\end{itemize}

\paragraph{Opinion Summary.} We evaluate the output performance according to the following dimensions\footnote{Unlike background summaries, opinion summaries are inherently \textit{integrative}. The space of opinions that could be wrongly integrated is effectively unbounded, making it infeasible to construct an exhaustive set of all incorrect atomic opinions to be excluded. Therefore, measuring a \textbf{Focus} dimension is impractical. Instead, we identify integration errors in the summary for uncovered clear atomic opinions.}:
\begin{itemize}[leftmargin=12pt, itemsep=0pt, topsep=2pt]
  \item \textbf{Overall Quality}: Whether the summary successfully integrates clarified opinions, enabling it to cover all clear atomic opinions.
  \item \textbf{Integration Error}: For uncovered clear atomic opinions, identify the errors in the summary.
\end{itemize}

\paragraph{Summarization Paradigm.} We assess the overall quality of each summarization paradigm:
\begin{itemize}[leftmargin=12pt, itemsep=0pt, topsep=2pt]
  \item \textbf{Overall Quality}: How well the background and opinion summaries work together.
\end{itemize}

\subsection{Methods and Metrics}
\label{sec:methods_and_metrics}
\noindent{\textbf{EBS Evaluation.}} \quad We evaluate its quality by matching the paragraph indices from the system output against our annotations of supporting and nonsupporting paragraphs. We utilize \textbf{Supporting Paragraph Recall, Precision, and F1-score} to measure the coverage, focus, and overall quality of the summary, respectively.

\vspace{7pt}
\noindent{\textbf{ABS Evaluation.}} \quad We employ an LLM-based verifier \cite{wei2024longformfactualitylargelanguage} to check if the summary entails our annotated atomic facts. This process determines whether the summary includes key supporting facts while excluding nonsupporting ones. We use three fine-grained metrics, \textbf{Key Supporting Atomic Fact Recall, Precision, and F1-score}, to quantify the three dimensions of the summary. We provide detailed formulas in Appendix~\ref{sec:abs_evaluation_metrics}.

\vspace{7pt}
\noindent{\textbf{AOS Evaluation.}} \enspace Similar to ABS, we utilize an opinion verifier to measure its coverage of our annotated atomic opinions. We utilize \textbf{Clear Atomic Opinion Recall} to assess its overall quality. For each uncovered atomic opinion, we perform LLM-based error detection to identify the specific integration failure. Specifically, we define five fine-grained error types: implicit reference unclarified, implicit reference incorrectly clarified, opinion misattribution, opinion fact inconsistency, and opinion sentiment distortion. Detailed formulas and error definitions are provided in Appendix~\ref{sec:aos_evaluation_metrics} and \ref{sec:error_detection_instruction}.

\vspace{7pt}
\noindent{\textbf{Paradigm-Level Quality Evaluation.}} \enspace A successful KGDS output requires high quality in both its background and opinion summaries. To capture this dependency, we evaluate the paradigm-level quality using the \textbf{Geometric Mean} of the two sub-summary overall scores. The multiplicative nature of this metric ensures a high score is only achieved when both components are strong, reflecting their joint fulfillment of the task goal. Moreover, root normalization maintains dimensional consistency between the paradigm-level score and its parts.

\begin{table*}[t]
    \centering
    \small
    \setlength{\abovecaptionskip}{5pt}
    \setlength{\tabcolsep}{5pt}
    \resizebox{\textwidth}{!}{
    \begin{tabular}{lcccccc}
        \toprule[1pt]
        \multirow{4.2}{*}{\textbf{Model Name}} & 
        \multicolumn{6}{c}{\textbf{KGDS BenchMark} (\textit{single-turn structured-prompt and multi-turn self-reflection})} \\
        \cmidrule(lr){2-7}
        & \multicolumn{3}{c}{\textbf{EBS-AOS Paradigm}} & 
        \multicolumn{3}{c}{\textbf{ABS-AOS Paradigm}} \\
        \cmidrule(lr){2-4}
        \cmidrule(lr){5-7}
        & \textbf{SP} (\textbf{R-P-F1}) & \textbf{CAO} (\textbf{R}) & \textbf{PLQ} (\textbf{GM}) & \textbf{KSAF} (\textbf{R-P-F1}) & \textbf{CAO} (\textbf{R}) & \textbf{PLQ} (\textbf{GM}) \\
        \midrule[0.5pt]
        GPT-4o & \multicolumn{1}{|c}{73.39\textcolor{green}{$_{\uparrow 1.56}$} 88.10\textcolor{green}{$_{\uparrow 0.34}$} \textbf{78.12}\textcolor{green}{$_{\uparrow 1.27}$}} & \multicolumn{1}{|c}{\textbf{76.18}\textcolor{green}{$_{\uparrow 0.54}$}} & \multicolumn{1}{|c}{\textbf{76.24}\textcolor{green}{$_{\uparrow 1.04}$}} & \multicolumn{1}{|c}{63.57\textcolor{red}{$_{\downarrow 1.08}$} 61.73\textcolor{green}{$_{\uparrow 1.62}$} \textbf{58.34}\textcolor{green}{$_{\uparrow 0.60}$}} & \multicolumn{1}{|c}{\textbf{69.42}\textcolor{red}{$_{\downarrow 0.21}$}} & \multicolumn{1}{|c}{\textbf{61.09}\textcolor{green}{$_{\uparrow 0.47}$}} \\
        GPT-4-turbo & \multicolumn{1}{|c}{73.03\textcolor{red}{$_{\downarrow 7.72}$} 87.42\textcolor{green}{$_{\uparrow 1.67}$} 77.14\textcolor{red}{$_{\downarrow 3.80}$}} & \multicolumn{1}{|c}{72.73\textcolor{green}{$_{\uparrow 0.45}$}} & \multicolumn{1}{|c}{73.94\textcolor{red}{$_{\downarrow 1.47}$}} & \multicolumn{1}{|c}{46.28\textcolor{red}{$_{\downarrow 5.78}$} 54.42\textcolor{green}{$_{\uparrow 5.33}$} 47.26\textcolor{red}{$_{\downarrow 1.64}$}} & \multicolumn{1}{|c}{58.32\textcolor{red}{$_{\downarrow 1.90}$}} & \multicolumn{1}{|c}{48.28\textcolor{red}{$_{\downarrow 0.74}$}} \\
        GPT-4o-mini & \multicolumn{1}{|c}{76.23\textcolor{red}{$_{\downarrow 0.30}$} 67.71\textcolor{green}{$_{\uparrow 0.31}$} 67.92\textcolor{red}{$_{\downarrow 0.03}$}} & \multicolumn{1}{|c}{29.08\textcolor{red}{$_{\downarrow 1.00}$}} & \multicolumn{1}{|c}{34.24\textcolor{red}{$_{\downarrow 0.97}$}} & \multicolumn{1}{|c}{33.75\textcolor{green}{$_{\uparrow 0.48}$} 45.07\textcolor{green}{$_{\uparrow 0.52}$} 36.51\textcolor{green}{$_{\uparrow 0.52}$}} & \multicolumn{1}{|c}{27.74\textcolor{red}{$_{\downarrow 0.41}$}} & \multicolumn{1}{|c}{24.64\textcolor{red}{$_{\downarrow 0.21}$}} \\
        \midrule[0.5pt]
        Claude 3 Opus & \multicolumn{1}{|c}{65.60\textcolor{green}{$_{\uparrow 2.21}$} 85.01\textcolor{red}{$_{\downarrow 4.57}$} 72.07\textcolor{red}{$_{\downarrow 1.31}$}} & \multicolumn{1}{|c}{\textbf{75.20}\textcolor{red}{$_{\downarrow 2.19}$}} & \multicolumn{1}{|c}{72.09\textcolor{red}{$_{\downarrow 2.45}$}} & \multicolumn{1}{|c}{51.10\textcolor{green}{$_{\uparrow 1.03}$} 74.59\textcolor{red}{$_{\downarrow 1.13}$} \textbf{58.03}\textcolor{red}{$_{\downarrow 0.91}$}} & \multicolumn{1}{|c}{\textbf{69.95}\textcolor{red}{$_{\downarrow 2.41}$}} & \multicolumn{1}{|c}{\textbf{60.72}\textcolor{red}{$_{\downarrow 1.29}$}} \\
        Claude 3.5 Sonnet & \multicolumn{1}{|c}{80.36\textcolor{green}{$_{\uparrow 2.42}$} 87.93\textcolor{red}{$_{\downarrow 0.62}$} \textbf{82.33}\textcolor{green}{$_{\uparrow 0.46}$}} & \multicolumn{1}{|c}{74.93\textcolor{red}{$_{\downarrow 5.30}$}} & \multicolumn{1}{|c}{\textbf{77.12}\textcolor{red}{$_{\downarrow 4.45}$}} & \multicolumn{1}{|c}{40.74\textcolor{green}{$_{\uparrow 5.84}$} 65.18\textcolor{red}{$_{\downarrow 0.79}$} 47.75\textcolor{green}{$_{\uparrow 3.51}$}} & \multicolumn{1}{|c}{58.55\textcolor{green}{$_{\uparrow 1.29}$}} & \multicolumn{1}{|c}{48.83\textcolor{green}{$_{\uparrow 2.60}$}} \\
        Claude 3.5 Haiku & \multicolumn{1}{|c}{62.69\textcolor{red}{$_{\downarrow 8.39}$} 76.01\textcolor{green}{$_{\uparrow 0.37}$} 66.02\textcolor{red}{$_{\downarrow 5.66}$}} & \multicolumn{1}{|c}{40.35\textcolor{red}{$_{\downarrow 9.13}$}} & \multicolumn{1}{|c}{46.02\textcolor{red}{$_{\downarrow 10.1}$}} & \multicolumn{1}{|c}{37.42\textcolor{red}{$_{\downarrow 6.90}$} 48.72\textcolor{red}{$_{\downarrow 4.65}$} 39.91\textcolor{red}{$_{\downarrow 6.63}$}} & \multicolumn{1}{|c}{29.64\textcolor{red}{$_{\downarrow 5.71}$}} & \multicolumn{1}{|c}{27.22\textcolor{red}{$_{\downarrow 4.48}$}} \\
        \midrule[0.5pt]
        Gemini 1.5 Pro & \multicolumn{1}{|c}{84.64\textcolor{red}{$_{\downarrow 4.38}$} 82.25\textcolor{green}{$_{\uparrow 4.43}$} \textbf{79.73}\textcolor{green}{$_{\uparrow 1.36}$}} & \multicolumn{1}{|c}{\textbf{76.86}\textcolor{red}{$_{\downarrow 0.53}$}} & \multicolumn{1}{|c}{\textbf{76.71}\textcolor{green}{$_{\uparrow 0.49}$}} & \multicolumn{1}{|c}{50.40\textcolor{red}{$_{\downarrow 4.13}$} 52.33\textcolor{green}{$_{\uparrow 0.28}$} 48.42\textcolor{red}{$_{\downarrow 2.11}$}} & \multicolumn{1}{|c}{\textbf{69.09}\textcolor{red}{$_{\downarrow 6.08}$}} & \multicolumn{1}{|c}{\textbf{54.83}\textcolor{red}{$_{\downarrow 4.45}$}} \\
        Llama-3.1-405B & \multicolumn{1}{|c}{79.09\textcolor{green}{$_{\uparrow 2.72}$} 77.96\textcolor{red}{$_{\downarrow 1.60}$} 75.16\textcolor{green}{$_{\uparrow 0.36}$}} & \multicolumn{1}{|c}{64.25\textcolor{red}{$_{\downarrow 0.01}$}} & \multicolumn{1}{|c}{67.58\textcolor{green}{$_{\uparrow 0.01}$}} & \multicolumn{1}{|c}{38.19\textcolor{red}{$_{\downarrow 1.08}$} 58.47\textcolor{green}{$_{\uparrow 7.38}$} 43.39\textcolor{green}{$_{\uparrow 1.12}$}} & \multicolumn{1}{|c}{52.98\textcolor{red}{$_{\downarrow 2.44}$}} & \multicolumn{1}{|c}{43.79\textcolor{green}{$_{\uparrow 0.11}$}} \\
        Mistral Large & \multicolumn{1}{|c}{68.81\textcolor{green}{$_{\uparrow 1.40}$} 78.55\textcolor{red}{$_{\downarrow 0.18}$} 71.07\textcolor{green}{$_{\uparrow 0.89}$}} & \multicolumn{1}{|c}{60.82\textcolor{red}{$_{\downarrow 2.62}$}} & \multicolumn{1}{|c}{63.63\textcolor{red}{$_{\downarrow 1.59}$}} & \multicolumn{1}{|c}{53.91\textcolor{red}{$_{\downarrow 3.05}$} 56.78\textcolor{red}{$_{\downarrow 0.32}$} \textbf{52.57}\textcolor{red}{$_{\downarrow 1.94}$}} & \multicolumn{1}{|c}{46.24\textcolor{red}{$_{\downarrow 1.71}$}} &\multicolumn{1}{|c}{46.36\textcolor{red}{$_{\downarrow 2.26}$}} \\
        \midrule[0.5pt]
        DeepSeek-V3 & \multicolumn{1}{|c}{86.98\textcolor{red}{$_{\downarrow 1.19}$} 73.83\textcolor{green}{$_{\uparrow 1.37}$} 75.66\textcolor{green}{$_{\uparrow 1.07}$}} & \multicolumn{1}{|c}{64.98\textcolor{green}{$_{\uparrow 1.84}$}} & \multicolumn{1}{|c}{66.50\textcolor{green}{$_{\uparrow 1.61}$}} & \multicolumn{1}{|c}{47.64\textcolor{green}{$_{\uparrow 0.65}$} 42.12\textcolor{green}{$_{\uparrow 0.60}$} 42.17\textcolor{green}{$_{\uparrow 0.65}$}} & \multicolumn{1}{|c}{56.02\textcolor{green}{$_{\uparrow 0.77}$}} & \multicolumn{1}{|c}{44.09\textcolor{green}{$_{\uparrow 0.72}$}} \\
        Qwen-Max & \multicolumn{1}{|c}{74.59\textcolor{red}{$_{\downarrow 5.53}$} 79.86\textcolor{red}{$_{\downarrow 1.30}$} 73.79\textcolor{red}{$_{\downarrow 2.78}$}} & \multicolumn{1}{|c}{60.93\textcolor{red}{$_{\downarrow 1.24}$}} & \multicolumn{1}{|c}{64.34\textcolor{red}{$_{\downarrow 1.59}$}} & \multicolumn{1}{|c}{46.49\textcolor{red}{$_{\downarrow 0.02}$} 53.59\textcolor{red}{$_{\downarrow 0.35}$} 46.87\textcolor{red}{$_{\downarrow 0.17}$}} & \multicolumn{1}{|c}{45.29\textcolor{red}{$_{\downarrow 1.54}$}} & \multicolumn{1}{|c}{40.87\textcolor{red}{$_{\downarrow 0.23}$}} \\
        GLM-4-Plus & \multicolumn{1}{|c}{80.28\textcolor{red}{$_{\downarrow 1.03}$} 72.81\textcolor{green}{$_{\uparrow 0.61}$} 71.17\textcolor{green}{$_{\uparrow 0.34}$}} & \multicolumn{1}{|c}{69.34\textcolor{red}{$_{\downarrow 1.73}$}} & \multicolumn{1}{|c}{67.55\textcolor{red}{$_{\downarrow 0.43}$}} & \multicolumn{1}{|c}{43.50\textcolor{red}{$_{\downarrow 1.77}$} 46.93\textcolor{green}{$_{\uparrow 0.43}$} 41.64\textcolor{red}{$_{\downarrow 0.61}$}} & \multicolumn{1}{|c}{55.80\textcolor{red}{$_{\downarrow 3.17}$}} & \multicolumn{1}{|c}{43.54\textcolor{red}{$_{\downarrow 1.26}$}} \\
        \bottomrule[1pt]
    \end{tabular}
    }
    \caption{Main evaluation results. \textbf{SP}, \textbf{KSAF}, \textbf{CAO}, and \textbf{PLQ} represent Supporting Paragraph, Key Supporting Atomic Fact, Clear Atomic Opinion, and Paradigm-Level Quality, respectively. \textbf{R}, \textbf{P}, \textbf{F1}, and \textbf{GM} denote Recall, Precision, F1-score, and Geometric Mean, respectively. All reported metrics are macro-averaged (\scalebox{0.90}{$\%$}). \textcolor{green}{$\uparrow$} and \textcolor{red}{$\downarrow$} indicate performance \textcolor{green}{\textbf{increases}} and \textcolor{red}{\textbf{decreases}}, respectively, after self-reflection following the structured-prompt. For each overall metric (i.e., $\mathrm{SP}_{\mathrm{F_1}}$, $\mathrm{KSAF}_{\mathrm{F_1}}$, $\mathrm{CAO}_{\mathrm{R}}$, and $\mathrm{PLQ}_{\mathrm{GM}}$), we highlight the top-3 performing models.}
    \label{tab:main_evaluation_results}
    \vspace{-10pt}
\end{table*}

\section{Experimental Setup}
\paragraph{System Selection.} We select 12 LLMs as summarizers, covering the most advanced and lightweight variants\footnote{Our evaluation began on \textit{January}, 2025, and all LLMs used the latest official API versions available at that time.}: \textbf{GPT-4o}, \textbf{GPT-4-turbo}, \textbf{GPT-4o-mini}, \textbf{Claude 3 Opus}, \textbf{Claude 3.5 Sonnet}, \textbf{Claude 3.5 Haiku}, \textbf{Gemini 1.5 Pro}, \textbf{Llama-3.1-405B}, \textbf{Mistral Large}, \textbf{DeepSeek-V3}, \textbf{Qwen-Max}, \textbf{GLM-4-Plus}. All model sources are listed in Appendix~\ref{sec:llm_sources}.

\paragraph{Prompt Engineering.} We benchmark LLMs for KGDS under two interaction patterns: (1) Single-turn \textbf{Structured-prompt}: A well-structured standard prompt \cite{li2023structuredchainofthoughtpromptingcode} that includes input content, input definition, task description, output definition, and return format (Appendix~\ref{sec:structured_prompts}). (2) Multi-turn \textbf{Self-reflection}: A second-round self-reflection instruction \cite{NEURIPS2023_1b44b878} with step-by-step chain-of-thought reasoning \cite{NEURIPS2022_9d560961} following the structured prompt (Appendix~\ref{sec:self-reflection_instructions}).

\paragraph{Verifier and Detector.} We use GPT-4o\footnote{\texttt{gpt-4o-2024-11-20}} to perform fact and opinion verification as well as error detection due to its excellent consistency with human judgment \cite{song-etal-2024-finesure, song-etal-2024-veriscore}. All prompts are provided in Appendices~\ref{sec:verification_prompts} and~\ref{sec:error_detection_instruction}.

\begin{figure}[t]
  \centering
  \setlength{\abovecaptionskip}{0pt}
  \includegraphics[width=\linewidth]{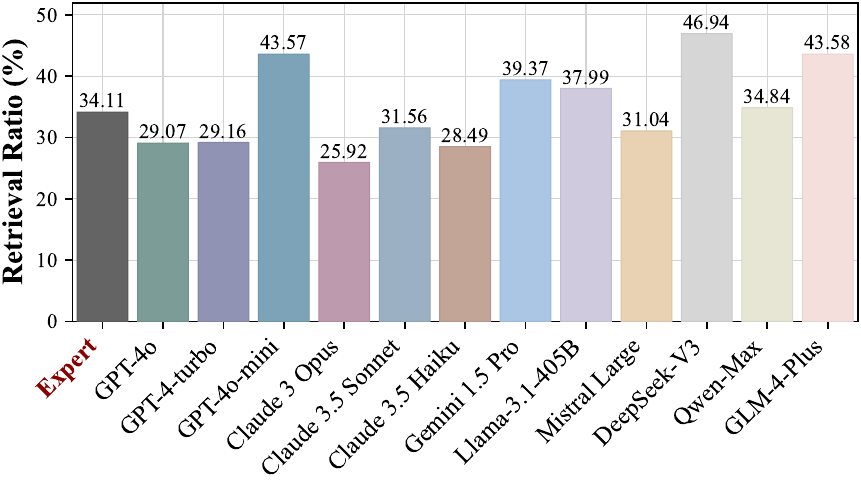}
  \caption{Paragraph retrieval ratios (\scalebox{0.90}{$\%$}) of LLMs. The majority of models can be classified as either conservative ($ratio<30\scalebox{0.90}{$\%$}$) or open retrievers ($ratio>38\scalebox{0.90}{$\%$}$).}
  \label{fig:paragraph_retrieval_ratios}
  \vspace{-10pt}
\end{figure}

\section{Analysis}
\vspace{-2pt}
In this section, we primarily reveal the challenges LLMs face in KGDS and analyze the commonalities and differences among them. Sections~\S\ref{sec:background_summary_analysis}, \S\ref{sec:opinion_summary_analysis}, and \S\ref{sec:overall_performance_analysis} respectively discuss the performance in background summary, opinion summary, and both paradigms under the structured-prompt setting. \S\ref{sec:self_reflection_impact} explores the impact of LLM self-reflection.

\vspace{-1pt}
\subsection{Background Summary}
\label{sec:background_summary_analysis}
\paragraph{{LLMs are moderate but imbalanced retrievers for EBS.}} From Table~\ref{tab:main_evaluation_results}, we find that most LLMs achieve moderate retrieval performance ($\mathrm{SP}_{\mathrm{F_1}}\in[71, 82]$) and lightweight models (i.e., GPT-4o-mini and Claude 3.5 Haiku) perform poorly. However, from $\mathrm{SP}_{\mathrm{R}}$ and $\mathrm{SP}_{\mathrm{P}}$, we observe significant polarization among LLMs, which indicates distinct retrieval strategies: some prioritize precision at the cost of recall (e.g., GPT-4-turbo), while others do the exact opposite (e.g., DeepSeek-V3). Such imbalance reveals that the current LLMs struggle with the \textit{precision-recall trade-off}. Moreover, we investigate the paragraph retrieval ratio\footnote{Defined as the ratio of the number of retrieved paragraphs to the total number of paragraphs.} (Figure~\ref{fig:paragraph_retrieval_ratios}) and identify that most LLMs exhibit either under- or over-retrieval, which is consistent with imbalance.

\vspace{-2pt}
\paragraph{LLMs are inadequate generators for ABS.} As presented in Table~\ref{tab:main_evaluation_results}, all LLMs exhibit unsatisfactory performance ($\mathrm{KSAF}_{\mathrm{F_1}}\in[37, 58]$). The low $\mathrm{KSAF}_{\mathrm{R}}$ (average of 46.08\%) reveals that LLMs often omit key facts, while the weak $\mathrm{KSAF}_{\mathrm{P}}$ (average of 54.99\%) indicates persistent inclusion of irrelevant facts. This dual-failure reflects the fundamental deficiencies of LLMs in meeting the requirements of \textit{coverage} and \textit{focus}. We also observe polarization among LLMs: some prioritize precision at the cost of recall (e.g., Claude 3.5 Sonnet), while others attempt to balance both (e.g., GPT-4o). Unlike EBS, we do not find any extreme recall-oriented models, indicating that LLMs tend to be either conservative or balanced in ABS generation.

\subsection{Opinion Summary}
\label{sec:opinion_summary_analysis}
\begin{figure}[t]
  \centering
  \setlength{\abovecaptionskip}{3pt}
  \includegraphics[width=\linewidth]{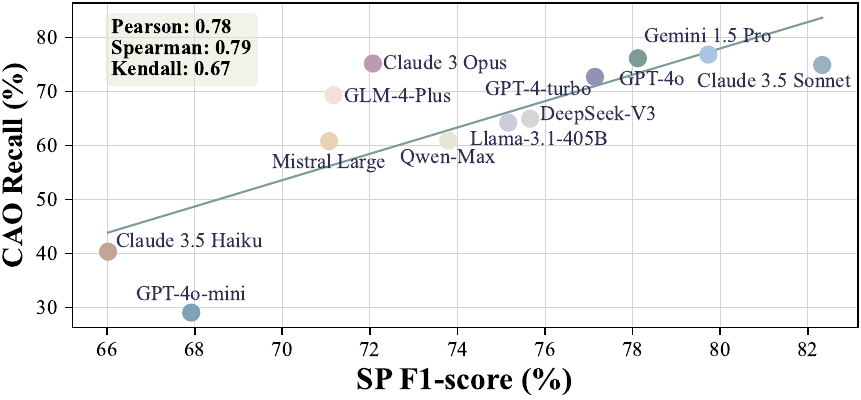}
  \caption{Visualization and metrics of the correlation between $\mathrm{SP}_{\mathrm{F_1}}$ and $\mathrm{CAO}_{\mathrm{R}}$ under the EBS-AOS paradigm.}
  \label{fig:bsp_cao_rela}
  \vspace{-10pt}
\end{figure}

\begin{figure}[t]
  \centering
  \setlength{\abovecaptionskip}{3pt}
  \includegraphics[width=\linewidth]{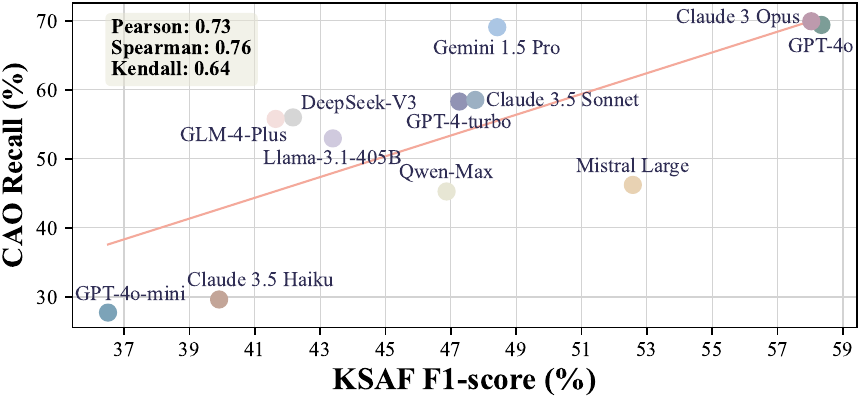}
  \caption{Vis. and metrics of the correlation between $\mathrm{KSAF}_{\mathrm{F_1}}$ and $\mathrm{CAO}_{\mathrm{R}}$ under the ABS-AOS paradigm.}
  \label{fig:kbsaf_cao_rela}
  \vspace{-1pt}
\end{figure}

\vspace{1pt}
\paragraph{Investigating correlation variables influencing AOS quality.} From Table~\ref{tab:main_evaluation_results}, we find that $\mathrm{CAO}_{\mathrm{R}}$ decreases as the background summary quality declines (i.e., $\mathrm{SP}_{\mathrm{F_1}}\rightarrow\mathrm{KSAF}_{\mathrm{F_1}}$) across all LLMs when switching paradigms. Furthermore, as shown in Figures~\ref{fig:bsp_cao_rela} and \ref{fig:kbsaf_cao_rela}, $\mathrm{CAO}_{\mathrm{R}}$ is highly positively correlated with $\mathrm{SP}_{\mathrm{F_1}}$ and $\mathrm{KSAF}_{\mathrm{F_1}}$ among LLMs in both independent paradigms. These findings indicate that \textit{a high-quality background summary facilitates opinion integration}. Meanwhile, individual differences suggest that the \textit{model-inherent integration ability is also a key factor} affecting AOS performance. For instance, Claude 3.5 Sonnet and Gemini 1.5 Pro show relatively weaker (i.e., below the line in Figure~\ref{fig:bsp_cao_rela}) and stronger (i.e., above the line in Figure~\ref{fig:kbsaf_cao_rela}) integration capabilities in EBS-AOS and ABS-AOS paradigms, respectively.

Nevertheless, even the most advanced models achieve only an average of 72\% $\mathrm{CAO}_{\mathrm{R}}$ across both paradigms, indicating that numerous opinions are being incorrectly integrated. Moreover, lightweight models (i.e., GPT-4o-mini and Claude 3.5 Haiku) exhibit significant performance flaws, suggesting that opinion integration is sensitive to model scale.

\paragraph{AOS error distribution analysis.} As illustrated in Table~\ref{tab:error_detection_results}, a significant ratio of errors is concentrated in IRU and IRIC across all LLMs under both paradigms, indicating that LLMs \textit{struggle to clarify implicit references effectively and correctly} during opinion integration. Moreover, although the ratios of OFI, OSD, and OM errors are relatively lower, their presence still highlights inherent deficiencies such as factual inconsistency, sentiment distortion, and incorrect participant assignment in LLMs.

\subsection{Summarization Paradigm}
\label{sec:overall_performance_analysis}
\paragraph{Performance stratification among LLMs.} We observe that the overall metric $\mathrm{PLQ}_{\mathrm{GM}}$ exhibits distinct hierarchies in independent paradigms and cross-paradigm (Figures~\ref{fig:op_ebs-aos}, \ref{fig:op_abs-aos}, and \ref{fig:op_avg-patterns} in Appendix~\ref{sec:overall_performance_visualization}). This suggests that the performance of LLMs improves in a stepwise manner rather than continuously as their intelligence advances in KGDS. Nevertheless, even the best-performing models achieve less than 69\% average performance across both paradigms (see Figure~\ref{fig:op_avg-patterns}), highlighting that KGDS remains a challenging task for current LLMs.

\paragraph{Our evaluation framework aligns well with human judgment.} To validate this, we conduct a human evaluation of overall paradigm-level quality\footnote{We focus on paradigm-level validation because the KGDS task's success depends on the complementarity of both sub-summaries. Furthermore, the individual sub-summary evaluations are already grounded in expert annotations and interpretable, fine-grained verification methods that are known to correlate well with human assessment \cite{song-etal-2024-finesure}.}. We randomly sample 120 summary pairs (10 per model and 5 per paradigm) from all outputs, and two human evaluators rate their overall quality on a 1-5 Likert scale \cite{joshi2015likert}, using the average as the final human score. For comparison, we select three LLMs (i.e., GPT-4o, Llama-3.1-405B, DeepSeek-V3) as LLM-as-a-judge baselines to directly score the summaries, with prompt settings following G-Eval \cite{liu-etal-2023-g}. Table~\ref{tab:evaluation_framework} presents the Pearson, Spearman, and Kendall correlation coefficients. Our framework achieves significantly higher correlation with human judgments than the baselines, demonstrating the effectiveness of our fine-grained, hierarchical approach.

\begin{table}[t]
    \centering
    \small
    \setlength{\abovecaptionskip}{5pt}
    \setlength{\tabcolsep}{5pt}
    \begin{tabular}{lccc}
        \toprule[1pt]
        \textbf{Metric} & \textbf{Pearson} & \textbf{Spearman} & \textbf{Kendall} \\
        \midrule[0.5pt]
        \rowcolor{gray!20}
        \multicolumn{4}{l}{\textit{LLM-as-a-judge}} \\
        GPT-4o & .5273 & .5162 & .4193 \\
        Llama-3.1-405B & .3506 & .3760 & .2881 \\
        DeepSeek-V3 & .4282 & .4417 & .3479 \\
        \midrule[0.5pt]
        \textbf{Our Framework} & \textbf{.6717} & \textbf{.6602} & \textbf{.5829} \\
        \bottomrule[1pt]
    \end{tabular}
    \caption{Correlation coefficients between different evaluation methods and human judgments, where our framework demonstrates the best alignment ($p < 0.01$).}
    \label{tab:evaluation_framework}
    \vspace{-14pt}
\end{table}

\begin{table*}[t]
    \centering
    \small
    \setlength{\abovecaptionskip}{5pt}
    \setlength{\tabcolsep}{5pt}
    \resizebox{\textwidth}{!}{
    \begin{tabular}{lcccccccccc}
        \toprule[1pt]
        \multirow{4.2}{*}{\textbf{Model Name}} & 
        \multicolumn{10}{c}{\textbf{AOS Error Detection} (\textit{single-turn structured-prompt and multi-turn self-reflection})} \\
        \cmidrule(lr){2-11}
        & \multicolumn{5}{c}{\textbf{EBS-AOS Paradigm}} & 
        \multicolumn{5}{c}{\textbf{ABS-AOS Paradigm}} \\
        \cmidrule(lr){2-6}
        \cmidrule(lr){7-11}
        & \textbf{OFI} & \textbf{OSD} & \textbf{IRU} & \textbf{IRIC} & \textbf{OM} & \textbf{OFI} & \textbf{OSD} & \textbf{IRU} & \textbf{IRIC} & \textbf{OM} \\
        \midrule[0.5pt]
        GPT-4o & \multicolumn{1}{|c}{8.57\textcolor{red}{$_{\uparrow 1.09}$}} & 6.19\textcolor{green}{$_{\downarrow 0.88}$} & 26.67\textcolor{red}{$_{\uparrow 2.80}$} & 55.71\textcolor{green}{$_{\downarrow 2.09}$} & 2.86\textcolor{green}{$_{\downarrow 0.92}$} & \multicolumn{1}{|c}{4.07\textcolor{green}{$_{\downarrow 1.12}$}} & 2.96\textcolor{red}{$_{\uparrow 1.10}$} & 50.74\textcolor{green}{$_{\downarrow 1.66}$} & 40.37\textcolor{red}{$_{\uparrow 1.33}$} & 1.86\textcolor{red}{$_{\uparrow 0.35}$} \\
        GPT-4-turbo & \multicolumn{1}{|c}{7.23\textcolor{green}{$_{\downarrow 0.38}$}} & 9.24\textcolor{green}{$_{\downarrow 1.18}$} & 40.96\textcolor{red}{$_{\uparrow 1.38}$} & 39.76\textcolor{green}{$_{\downarrow 0.65}$} & 2.81\textcolor{red}{$_{\uparrow 0.83}$} & \multicolumn{1}{|c}{6.74\textcolor{red}{$_{\uparrow 0.40}$}} & 3.23\textcolor{red}{$_{\uparrow 2.33}$} & 56.33\textcolor{green}{$_{\downarrow 0.51}$} & 31.00\textcolor{green}{$_{\downarrow 1.11}$} & 2.70\textcolor{green}{$_{\downarrow 1.11}$} \\
        GPT-4o-mini & \multicolumn{1}{|c}{0.79\textcolor{red}{$_{\uparrow 0.30}$}} & 0.32\textcolor{red}{$_{\uparrow 0.15}$} & 81.80\textcolor{green}{$_{\downarrow 0.08}$} & 16.93\textcolor{green}{$_{\downarrow 0.37}$} & 0.16\textcolor{green}{$_{\downarrow 0.00}$} & \multicolumn{1}{|c}{1.74\textcolor{green}{$_{\downarrow 0.79}$}} & 0.79\textcolor{red}{$_{\uparrow 0.15}$} & 73.69\textcolor{green}{$_{\downarrow 0.30}$} & 23.61\textcolor{red}{$_{\uparrow 1.11}$} & 0.17\textcolor{green}{$_{\downarrow 0.17}$} \\
        \midrule[0.5pt]
        Claude 3 Opus & \multicolumn{1}{|c}{6.96\textcolor{red}{$_{\uparrow 0.54}$}} & 2.17\textcolor{green}{$_{\downarrow 0.09}$} & 32.17\textcolor{red}{$_{\uparrow 0.75}$} & 56.96\textcolor{green}{$_{\downarrow 1.96}$} & 1.74\textcolor{red}{$_{\uparrow 0.76}$} & \multicolumn{1}{|c}{6.61\textcolor{red}{$_{\uparrow 0.64}$}} & 1.18\textcolor{red}{$_{\uparrow 0.99}$} & 46.69\textcolor{red}{$_{\uparrow 0.05}$} & 43.19\textcolor{green}{$_{\downarrow 2.61}$} & 2.33\textcolor{red}{$_{\uparrow 0.93}$} \\
        Claude 3.5 Sonnet & \multicolumn{1}{|c}{6.67\textcolor{red}{$_{\uparrow 0.24}$}} & 5.78\textcolor{green}{$_{\downarrow 1.78}$} & 50.22\textcolor{red}{$_{\uparrow 2.51}$} & 35.11\textcolor{green}{$_{\downarrow 1.29}$} & 2.22\textcolor{red}{$_{\uparrow 0.32}$} & \multicolumn{1}{|c}{3.00\textcolor{red}{$_{\uparrow 3.13}$}} & 3.54\textcolor{green}{$_{\downarrow 0.20}$} & 61.85\textcolor{green}{$_{\downarrow 8.65}$} & 29.97\textcolor{red}{$_{\uparrow 5.13}$} & 1.64\textcolor{red}{$_{\uparrow 0.59}$} \\
        Claude 3.5 Haiku & \multicolumn{1}{|c}{6.81\textcolor{green}{$_{\downarrow 2.01}$}} & 4.09\textcolor{green}{$_{\downarrow 0.83}$} & 55.45\textcolor{red}{$_{\uparrow 4.07}$} & 32.10\textcolor{green}{$_{\downarrow 1.23}$} & 1.55\textcolor{green}{$_{\downarrow 0.00}$} & \multicolumn{1}{|c}{3.91\textcolor{green}{$_{\downarrow 0.10}$}} & 1.47\textcolor{green}{$_{\downarrow 0.41}$} & 70.52\textcolor{green}{$_{\downarrow 2.64}$} & 23.94\textcolor{red}{$_{\uparrow 3.31}$} & 0.16\textcolor{green}{$_{\downarrow 0.16}$} \\
        \midrule[0.5pt]
        Gemini 1.5 Pro & \multicolumn{1}{|c}{6.28\textcolor{green}{$_{\downarrow 0.91}$}} & 7.73\textcolor{red}{$_{\uparrow 2.03}$} & 37.68\textcolor{red}{$_{\uparrow 3.78}$} & 46.86\textcolor{green}{$_{\downarrow 3.45}$} & 1.45\textcolor{green}{$_{\downarrow 1.45}$} & \multicolumn{1}{|c}{3.62\textcolor{red}{$_{\uparrow 0.05}$}} & 6.52\textcolor{green}{$_{\downarrow 0.40}$} & 63.41\textcolor{red}{$_{\uparrow 4.17}$} & 26.09\textcolor{green}{$_{\downarrow 4.38}$} & 0.36\textcolor{red}{$_{\uparrow 0.56}$} \\
        Llama-3.1-405B & \multicolumn{1}{|c}{6.92\textcolor{green}{$_{\downarrow 1.00}$}} & 1.89\textcolor{red}{$_{\uparrow 1.23}$} & 55.35\textcolor{red}{$_{\uparrow 0.10}$} & 34.59\textcolor{green}{$_{\downarrow 0.95}$} & 1.25\textcolor{red}{$_{\uparrow 0.62}$} & \multicolumn{1}{|c}{3.83\textcolor{red}{$_{\uparrow 1.18}$}} & 0.47\textcolor{red}{$_{\uparrow 1.81}$} & 69.62\textcolor{red}{$_{\uparrow 2.13}$} & 25.36\textcolor{green}{$_{\downarrow 5.77}$} & 0.72\textcolor{red}{$_{\uparrow 0.65}$} \\
       	Mistral Large & \multicolumn{1}{|c}{6.67\textcolor{red}{$_{\uparrow 1.35}$}} & 3.33\textcolor{green}{$_{\downarrow 0.39}$} & 52.78\textcolor{green}{$_{\downarrow 0.91}$} & 36.94\textcolor{green}{$_{\downarrow 1.91}$} & 0.28\textcolor{red}{$_{\uparrow 1.86}$} & \multicolumn{1}{|c}{3.73\textcolor{green}{$_{\downarrow 0.12}$}} & 2.28\textcolor{green}{$_{\downarrow 0.48}$} & 64.52\textcolor{red}{$_{\uparrow 5.82}$} & 29.05\textcolor{green}{$_{\downarrow 5.60}$} & 0.42\textcolor{red}{$_{\uparrow 0.38}$} \\
        \midrule[0.5pt]
        DeepSeek-V3 & \multicolumn{1}{|c}{4.79\textcolor{red}{$_{\uparrow 1.61}$}} & 2.24\textcolor{green}{$_{\downarrow 0.22}$} & 53.35\textcolor{green}{$_{\downarrow 5.54}$} & 38.98\textcolor{red}{$_{\uparrow 4.12}$} & 0.64\textcolor{red}{$_{\uparrow 0.03}$} & \multicolumn{1}{|c}{5.99\textcolor{green}{$_{\downarrow 0.67}$}} & 2.08\textcolor{red}{$_{\uparrow 0.05}$} & 61.98\textcolor{green}{$_{\downarrow 2.94}$} & 28.91\textcolor{red}{$_{\uparrow 3.80}$} & 1.04\textcolor{green}{$_{\downarrow 0.24}$} \\
        Qwen-Max & \multicolumn{1}{|c}{7.54\textcolor{red}{$_{\uparrow 0.13}$}} & 2.90\textcolor{red}{$_{\uparrow 0.79}$} & 43.48\textcolor{red}{$_{\uparrow 1.12}$} & 45.22\textcolor{green}{$_{\downarrow 2.32}$} & 0.86\textcolor{red}{$_{\uparrow 0.28}$} & \multicolumn{1}{|c}{4.47\textcolor{red}{$_{\uparrow 0.05}$}} & 3.25\textcolor{green}{$_{\downarrow 0.30}$} & 65.45\textcolor{red}{$_{\uparrow 1.94}$} & 26.22\textcolor{green}{$_{\downarrow 1.66}$} & 0.61\textcolor{green}{$_{\downarrow 0.03}$} \\
       	GLM-4-Plus & \multicolumn{1}{|c}{10.94\textcolor{green}{$_{\downarrow 1.09}$}} & 3.02\textcolor{red}{$_{\uparrow 1.72}$} & 40.38\textcolor{green}{$_{\downarrow 5.71}$} & 44.53\textcolor{red}{$_{\uparrow 4.74}$} & 1.13\textcolor{red}{$_{\uparrow 0.34}$} & \multicolumn{1}{|c}{5.23\textcolor{red}{$_{\uparrow 2.60}$}} & 1.93\textcolor{red}{$_{\uparrow 1.99}$} & 55.65\textcolor{red}{$_{\uparrow 0.22}$} & 35.81\textcolor{green}{$_{\downarrow 4.74}$} & 1.38\textcolor{green}{$_{\downarrow 0.07}$} \\
        \bottomrule[1pt]
    \end{tabular}
    }
    \caption{Fine-grained error detection results. \textbf{OFI}, \textbf{OSD}, \textbf{IRU}, \textbf{IRIC}, and \textbf{OM} represent the five error types: Opinion Fact Inconsistency, Opinion Sentiment Distortion, Implicit Reference Unclarified, Implicit Reference Incorrectly Clarified, and Opinion Misattribution. All reported metrics are error proportions (\scalebox{0.90}{$\%$}) when clear atomic opinions are not covered by AOS. \textcolor{green}{$\downarrow$} and \textcolor{red}{$\uparrow$} respectively indicate \textcolor{green}{\textbf{decreases}} and \textcolor{red}{\textbf{increases}} in error ratios after self-reflection.}
    \label{tab:error_detection_results}
    \vspace{-10pt}
\end{table*}

\subsection{Self-Reflection Impact}
\label{sec:self_reflection_impact}
\vspace{1pt}
\paragraph{Self-reflection does not essentially affect the performance of LLMs on KGDS.} From Table~\ref{tab:main_evaluation_results}, we observe that the performance fluctuations (i.e., \textcolor{green}{${\uparrow}$} or \textcolor{red}{${\downarrow}$}) after self-reflection are limited (the maximum average fluctuation across all metrics is $2.30$ for $\mathrm{CAO}_{\mathrm{R\text{:ABS-AOS}}}$). This means that self-reflection does not fundamentally influence the capacities of LLMs in KGDS, revealing two key limitations: (1) LLMs lack sufficient self-evaluation abilities for KGDS. (2) The reflection strategies of LLMs struggle to provide excellent reasoning paths for KGDS.

\paragraph{Self-reflection makes LLMs more risk-averse for opinion summary.} As presented in Table~\ref{tab:error_detection_results}, most LLMs exhibit a decrease in IRIC and an increase in IRU (e.g., GPT-4o with IRIC \textcolor{green}{${\downarrow 2.09}$}, IRU \textcolor{red}{${\uparrow 2.80}$}). This suggests that self-reflection encourages LLMs to adopt more cautious strategies for clarifying implicit references — \textit{opting to leave them unclarified rather than risk incorrect clarification}. However, such conservatism ultimately harms the quality of opinion summaries, as evidenced by the widespread drop in $\mathrm{CAO}_{\mathrm{R}}$ in Table~\ref{tab:main_evaluation_results}.

\vspace{-1pt}
\section{Related Work}
\vspace{-2pt}
\noindent{\textbf{Dialogue Summarization.}} \enspace This task spans diverse scenarios, including online chats \cite{gliwa-etal-2019-samsum}, meetings \cite{hu-etal-2023-meetingbank}, and customer service \cite{lin-etal-2021-csds}. Existing methods, whether based on feature modeling \cite{chen-yang-2021-structure, fang-etal-2022-spoken, lin-etal-2022-roles}, pre-training \cite{zou-etal-2021-low, zhong2022dialoglm}, or LLMs \cite{wang-etal-2023-instructive, zhu-etal-2025-factual}, focus heavily on single-source dialogue inputs while neglecting shared background knowledge among participants. Moreover, current evaluation systems \cite{ramprasad2024analyzing, tang2024tofueval} also assume single-source inputs. In contrast, our KGDS focuses on multi-source inputs and complementary outputs.

\vspace{5pt}
\noindent{\textbf{Knowledge-Intensive Dialogue Response.}} \enspace This task aims to enhance the dialogue process by incorporating external knowledge sources \cite{NEURIPS2020_6b493230, chen2024benchmarking}. It differs from our KGDS in two key aspects. First, knowledge-intensive dialogue focuses on improving the conversational experience for participants \cite{wang-etal-2023-large, wang2024improving}, whereas KGDS is designed to generate clear summaries for external readers. Second, the former addresses knowledge gaps between participants, making the dialogue inherently \textit{knowledge-intensive} \cite{zhang-etal-2020-grounded}. In contrast, KGDS centers on discussions among participants with shared knowledge, resulting in \textit{background-sparse} conversations that rely heavily on implicit context.

\vspace{5pt}
\noindent{\textbf{Summarization Evaluation.}} \enspace Earlier similarity-based metrics like Rouge \cite{lin2004rouge}, BERTScore \cite{zhang2019bertscore}, and MoverScore \cite{zhao-etal-2019-moverscore} are often misaligned with human judgment. Recently, LLM-based evaluators \cite{chen-etal-2023-exploring-use, shen-etal-2023-large, fu-etal-2024-gptscore} have been used, yet they still lack interpretability. To address this, some studies \cite{song-etal-2024-finesure, lee-etal-2024-unisumeval, yang-etal-2024-fizz, scire-etal-2024-fenice} utilize atomic facts \cite{min-etal-2023-factscore} and LLM-based claim verification \cite{song-etal-2024-veriscore}, achieving more fine-grained evaluation. Among these, FineSurE \cite{song-etal-2024-finesure} is most relevant to our work. However, its metrics lack consistency in evaluation granularity. In contrast, our coverage and focus metrics are both measured at the atomic fact level, ensuring consistent granularity across dimensions.

\vspace{-1pt}
\section{Conclusion}
\vspace{-2pt}
In this study, we introduce the KGDS task, which aims to create observer-centric summaries by integrating shared background knowledge with discussions. We establish the first benchmark for KGDS and propose a task-aligned hierarchical evaluation framework with fine-grained, interpretable metrics. Our evaluation of 12 leading LLMs reveals significant challenges in background summary retrieval, generation, and opinion summary integration, with even the most advanced models achieving less than 69\% average performance across both paradigms.

\section*{Limitations}
Our study presents several limitations that merit discussion and future exploration:

\vspace{-1pt}
\paragraph{Domain Generalization.} Knowledge-grounded discussions and their confusing summaries are prevalent in both open-domain and private scenarios. Our benchmark is centered on open-domain news discussions, which may limit its applicability to private scenarios (\textit{e.g.,} internal meetings, medical consultations, legal debates, etc.). However, shared background knowledge and the related discussions in private contexts are challenging to obtain and are less common and representative than our open benchmark. Future work will focus on validating the performance of LLMs in handling KGDS across diverse private scenarios.

\vspace{-1pt}
\paragraph{Language Diversity.} Our current research is focused on English, creating a monolingual benchmark that may not fully address the challenges of multilingual KGDS. This limitation could impact model performance in languages with different linguistic structures or sociocultural contexts. Nevertheless, English is the most prevalent language in both academic research and real-world LLM applications, making it a reasonable starting point. We plan to explore multilingual and cross-lingual KGDS in future research.

\section*{Ethics Statement}
Our KGDS benchmark is developed with careful consideration of ethical implications. The news articles used as shared background knowledge are publicly accessible through Google News and have been carefully reviewed to ensure they do not contain sensitive or harmful information. Explicit consent was obtained from all expert participants involved in discussion construction and data annotation, with a clear explanation of the research purpose and data usage protocols. Our evaluation methodology prioritizes factual accuracy and opinion fidelity to minimize the risk of hallucination propagation in real-world applications. Potential misuse risks, such as generating misleading summaries through improper background-discussion combinations, are mitigated through technical safeguards in our released code and explicit usage guidelines. Researchers using our KGDS benchmark should adhere to responsible AI principles, especially when applying similar techniques to sensitive domains like healthcare or legal discussions.

\paragraph{Annotation and Evaluation Cost.} Our annotation process is multi-step, fine-grained, and highly complex. For each sample in our benchmark, the average annotation time is $2.6$ hours (including news reading and understanding, discussion construction, supporting paragraphs, key supporting atomic facts, nonsupporting atomic facts, and clear atomic opinions annotation). We pay each annotation expert a wage of \textdollar$9$ per hour (above the minimum wage standard), and the total annotation cost for the KGDS benchmark is \textdollar$4680$ ($2.6$ hours per sample $\times$ \textdollar$9$ per hour $\times$ $2$ experts per sample $\times$ $100$ samples $=$ \textdollar$4680$). The evaluation cost of API calls is detailed in Table~\ref{tab:evaluation-cost}.

\begin{table}[htbp]
    \centering
    \small
    \setlength{\abovecaptionskip}{8pt}
    \setlength{\tabcolsep}{6pt}
    \begin{tabular}{lc}
        \toprule[1pt]
        \textbf{Project} & \multicolumn{1}{|c}{\textbf{Cost} (\textbf{\textdollar})} \\
        \midrule[0.5pt]
        Atomic Fact Decomposition & \multicolumn{1}{|c}{$21$} \\
        Conflicting Fact Masking & \multicolumn{1}{|c}{$16$} \\
        Structured Prompt & \multicolumn{1}{|c}{$74$} \\
        Reflection Instruction & \multicolumn{1}{|c}{$126$} \\
        Atomic Fact Verification & \multicolumn{1}{|c}{$173$} \\
        Atomic Opinion Verification & \multicolumn{1}{|c}{$158$} \\
        Fine-Grained Error Detection & \multicolumn{1}{|c}{$89$} \\
        \midrule[0.5pt]
        \textbf{Total} & \multicolumn{1}{|c}{$657$} \\
        \bottomrule[1pt]
    \end{tabular}
    \caption{API call costs for the KGDS benchmark.}
    \label{tab:evaluation-cost}
    \vspace{-12pt}
\end{table}

\section*{Acknowledgement}
This research was supported in part by the National Natural Science Foundation of China (Grant Nos. 62276017, 62406033, U1636211, 61672081) and the State Key Laboratory of Complex \& Critical Software Environment (Grant No. SKLCCSE-2024ZX-18). We thank the anonymous reviewers for their feedback that helped improve this work.

\vspace{-11pt}
\bibliography{custom}

@article{jia2023taxonomy,
  title={Taxonomy of Abstractive Dialogue Summarization: Scenarios, Approaches, and Future Directions},
  author={Jia, Qi and Liu, Yizhu and Ren, Siyu and Zhu, Kenny Q},
  journal={ACM Computing Surveys},
  volume={56},
  number={3},
  pages={1--38},
  year={2023},
  publisher={ACM New York, NY}
}

@inproceedings{yang-etal-2024-fizz,
    title = "{FIZZ}: Factual Inconsistency Detection by Zoom-in Summary and Zoom-out Document",
    author = "Yang, Joonho  and
      Yoon, Seunghyun  and
      Kim, ByeongJeong  and
      Lee, Hwanhee",
    editor = "Al-Onaizan, Yaser  and
      Bansal, Mohit  and
      Chen, Yun-Nung",
    booktitle = "Proceedings of the 2024 Conference on Empirical Methods in Natural Language Processing",
    month = nov,
    year = "2024",
    address = "Miami, Florida, USA",
    publisher = "Association for Computational Linguistics",
    url = "https://aclanthology.org/2024.emnlp-main.3/",
    doi = "10.18653/v1/2024.emnlp-main.3",
    pages = "30--45"
}

@inproceedings{shen-etal-2023-large,
    title = "Large Language Models are Not Yet Human-Level Evaluators for Abstractive Summarization",
    author = "Shen, Chenhui  and
      Cheng, Liying  and
      Nguyen, Xuan-Phi  and
      You, Yang  and
      Bing, Lidong",
    editor = "Bouamor, Houda  and
      Pino, Juan  and
      Bali, Kalika",
    booktitle = "Findings of the Association for Computational Linguistics: EMNLP 2023",
    month = dec,
    year = "2023",
    address = "Singapore",
    publisher = "Association for Computational Linguistics",
    url = "https://aclanthology.org/2023.findings-emnlp.278/",
    doi = "10.18653/v1/2023.findings-emnlp.278",
    pages = "4215--4233"
}

@misc{song2025learningsummarizellmgeneratedfeedback,
      title={Learning to Summarize from LLM-generated Feedback}, 
      author={Hwanjun Song and Taewon Yun and Yuho Lee and Jihwan Oh and Gihun Lee and Jason Cai and Hang Su},
      year={2025},
      eprint={2410.13116},
      archivePrefix={arXiv},
      primaryClass={cs.CL},
      url={https://arxiv.org/abs/2410.13116}, 
}

@inproceedings{fu-etal-2024-gptscore,
    title = "{GPTS}core: Evaluate as You Desire",
    author = "Fu, Jinlan  and
      Ng, See-Kiong  and
      Jiang, Zhengbao  and
      Liu, Pengfei",
    editor = "Duh, Kevin  and
      Gomez, Helena  and
      Bethard, Steven",
    booktitle = "Proceedings of the 2024 Conference of the North American Chapter of the Association for Computational Linguistics: Human Language Technologies (Volume 1: Long Papers)",
    month = jun,
    year = "2024",
    address = "Mexico City, Mexico",
    publisher = "Association for Computational Linguistics",
    url = "https://aclanthology.org/2024.naacl-long.365/",
    doi = "10.18653/v1/2024.naacl-long.365",
    pages = "6556--6576"
}

@inproceedings{scire-etal-2024-fenice,
    title = "{FENICE}: Factuality Evaluation of summarization based on Natural language Inference and Claim Extraction",
    author = "Scir{\`e}, Alessandro  and
      Ghonim, Karim  and
      Navigli, Roberto",
    editor = "Ku, Lun-Wei  and
      Martins, Andre  and
      Srikumar, Vivek",
    booktitle = "Findings of the Association for Computational Linguistics: ACL 2024",
    month = aug,
    year = "2024",
    address = "Bangkok, Thailand",
    publisher = "Association for Computational Linguistics",
    url = "https://aclanthology.org/2024.findings-acl.841/",
    doi = "10.18653/v1/2024.findings-acl.841",
    pages = "14148--14161"
}

@inproceedings{liu-etal-2023-g,
    title = "{G}-Eval: {NLG} Evaluation using Gpt-4 with Better Human Alignment",
    author = "Liu, Yang  and
      Iter, Dan  and
      Xu, Yichong  and
      Wang, Shuohang  and
      Xu, Ruochen  and
      Zhu, Chenguang",
    editor = "Bouamor, Houda  and
      Pino, Juan  and
      Bali, Kalika",
    booktitle = "Proceedings of the 2023 Conference on Empirical Methods in Natural Language Processing",
    month = dec,
    year = "2023",
    address = "Singapore",
    publisher = "Association for Computational Linguistics",
    url = "https://aclanthology.org/2023.emnlp-main.153/",
    doi = "10.18653/v1/2023.emnlp-main.153",
    pages = "2511--2522"
}

@inproceedings{zhao-etal-2019-moverscore,
    title = "{M}over{S}core: Text Generation Evaluating with Contextualized Embeddings and Earth Mover Distance",
    author = "Zhao, Wei  and
      Peyrard, Maxime  and
      Liu, Fei  and
      Gao, Yang  and
      Meyer, Christian M.  and
      Eger, Steffen",
    editor = "Inui, Kentaro  and
      Jiang, Jing  and
      Ng, Vincent  and
      Wan, Xiaojun",
    booktitle = "Proceedings of the 2019 Conference on Empirical Methods in Natural Language Processing and the 9th International Joint Conference on Natural Language Processing (EMNLP-IJCNLP)",
    month = nov,
    year = "2019",
    address = "Hong Kong, China",
    publisher = "Association for Computational Linguistics",
    url = "https://aclanthology.org/D19-1053/",
    doi = "10.18653/v1/D19-1053",
    pages = "563--578"
}

@article{zhang2019bertscore,
  title={Bertscore: Evaluating text generation with bert},
  author={Zhang, Tianyi and Kishore, Varsha and Wu, Felix and Weinberger, Kilian Q and Artzi, Yoav},
  journal={arXiv preprint arXiv:1904.09675},
  year={2019}
}

@article{rennard2023abstractive,
  title={Abstractive meeting summarization: A survey},
  author={Rennard, Virgile and Shang, Guokan and Hunter, Julie and Vazirgiannis, Michalis},
  journal={Transactions of the Association for Computational Linguistics},
  volume={11},
  pages={861--884},
  year={2023},
  publisher={MIT Press One Broadway, 12th Floor, Cambridge, Massachusetts 02142, USA~…}
}

@inproceedings{lin2004rouge,
  title={Rouge: A package for automatic evaluation of summaries},
  author={Lin, Chin-Yew},
  booktitle={Text summarization branches out},
  pages={74--81},
  year={2004}
}

@inproceedings{gliwa-etal-2019-samsum,
    title = "{SAMS}um Corpus: A Human-annotated Dialogue Dataset for Abstractive Summarization",
    author = "Gliwa, Bogdan  and
      Mochol, Iwona  and
      Biesek, Maciej  and
      Wawer, Aleksander",
    editor = "Wang, Lu  and
      Cheung, Jackie Chi Kit  and
      Carenini, Giuseppe  and
      Liu, Fei",
    booktitle = "Proceedings of the 2nd Workshop on New Frontiers in Summarization",
    month = nov,
    year = "2019",
    address = "Hong Kong, China",
    publisher = "Association for Computational Linguistics",
    url = "https://aclanthology.org/D19-5409",
    doi = "10.18653/v1/D19-5409",
    pages = "70--79",
}

@inproceedings{chen-etal-2021-dialogsum,
    title = "{D}ialog{S}um: {A} Real-Life Scenario Dialogue Summarization Dataset",
    author = "Chen, Yulong  and
      Liu, Yang  and
      Chen, Liang  and
      Zhang, Yue",
    editor = "Zong, Chengqing  and
      Xia, Fei  and
      Li, Wenjie  and
      Navigli, Roberto",
    booktitle = "Findings of the Association for Computational Linguistics: ACL-IJCNLP 2021",
    month = aug,
    year = "2021",
    address = "Online",
    publisher = "Association for Computational Linguistics",
    url = "https://aclanthology.org/2021.findings-acl.449",
    doi = "10.18653/v1/2021.findings-acl.449",
    pages = "5062--5074",
}

@inproceedings{zhu-etal-2021-mediasum,
    title = "{M}edia{S}um: A Large-scale Media Interview Dataset for Dialogue Summarization",
    author = "Zhu, Chenguang  and
      Liu, Yang  and
      Mei, Jie  and
      Zeng, Michael",
    editor = "Toutanova, Kristina  and
      Rumshisky, Anna  and
      Zettlemoyer, Luke  and
      Hakkani-Tur, Dilek  and
      Beltagy, Iz  and
      Bethard, Steven  and
      Cotterell, Ryan  and
      Chakraborty, Tanmoy  and
      Zhou, Yichao",
    booktitle = "Proceedings of the 2021 Conference of the North American Chapter of the Association for Computational Linguistics: Human Language Technologies",
    month = jun,
    year = "2021",
    address = "Online",
    publisher = "Association for Computational Linguistics",
    url = "https://aclanthology.org/2021.naacl-main.474",
    doi = "10.18653/v1/2021.naacl-main.474",
    pages = "5927--5934",
}

@inproceedings{hu-etal-2023-meetingbank,
    title = "{M}eeting{B}ank: A Benchmark Dataset for Meeting Summarization",
    author = "Hu, Yebowen  and
      Ganter, Timothy  and
      Deilamsalehy, Hanieh  and
      Dernoncourt, Franck  and
      Foroosh, Hassan  and
      Liu, Fei",
    editor = "Rogers, Anna  and
      Boyd-Graber, Jordan  and
      Okazaki, Naoaki",
    booktitle = "Proceedings of the 61st Annual Meeting of the Association for Computational Linguistics (Volume 1: Long Papers)",
    month = jul,
    year = "2023",
    address = "Toronto, Canada",
    publisher = "Association for Computational Linguistics",
    url = "https://aclanthology.org/2023.acl-long.906",
    doi = "10.18653/v1/2023.acl-long.906",
    pages = "16409--16423",
}

@inproceedings{lin-etal-2022-roles,
    title = "Other Roles Matter! Enhancing Role-Oriented Dialogue Summarization via Role Interactions",
    author = "Lin, Haitao  and
      Zhu, Junnan  and
      Xiang, Lu  and
      Zhou, Yu  and
      Zhang, Jiajun  and
      Zong, Chengqing",
    editor = "Muresan, Smaranda  and
      Nakov, Preslav  and
      Villavicencio, Aline",
    booktitle = "Proceedings of the 60th Annual Meeting of the Association for Computational Linguistics (Volume 1: Long Papers)",
    month = may,
    year = "2022",
    address = "Dublin, Ireland",
    publisher = "Association for Computational Linguistics",
    url = "https://aclanthology.org/2022.acl-long.182",
    doi = "10.18653/v1/2022.acl-long.182",
    pages = "2545--2558",
}

@inproceedings{zhou-etal-2023-multi,
    title = "Multi-Stage Pre-training Enhanced by {C}hat{GPT} for Multi-Scenario Multi-Domain Dialogue Summarization",
    author = "Zhou, Weixiao  and
      Li, Gengyao  and
      Cheng, Xianfu  and
      Liang, Xinnian  and
      Zhu, Junnan  and
      Zhai, Feifei  and
      Li, Zhoujun",
    editor = "Bouamor, Houda  and
      Pino, Juan  and
      Bali, Kalika",
    booktitle = "Findings of the Association for Computational Linguistics: EMNLP 2023",
    month = dec,
    year = "2023",
    address = "Singapore",
    publisher = "Association for Computational Linguistics",
    url = "https://aclanthology.org/2023.findings-emnlp.460",
    doi = "10.18653/v1/2023.findings-emnlp.460",
    pages = "6893--6908",
}

@inproceedings{wang-etal-2023-instructive,
    title = "Instructive Dialogue Summarization with Query Aggregations",
    author = "Wang, Bin  and
      Liu, Zhengyuan  and
      Chen, Nancy",
    editor = "Bouamor, Houda  and
      Pino, Juan  and
      Bali, Kalika",
    booktitle = "Proceedings of the 2023 Conference on Empirical Methods in Natural Language Processing",
    month = dec,
    year = "2023",
    address = "Singapore",
    publisher = "Association for Computational Linguistics",
    url = "https://aclanthology.org/2023.emnlp-main.474",
    doi = "10.18653/v1/2023.emnlp-main.474",
    pages = "7630--7653",
}

@inproceedings{gao-wan-2022-dialsummeval,
    title = "{D}ial{S}umm{E}val: Revisiting Summarization Evaluation for Dialogues",
    author = "Gao, Mingqi  and
      Wan, Xiaojun",
    editor = "Carpuat, Marine  and
      de Marneffe, Marie-Catherine  and
      Meza Ruiz, Ivan Vladimir",
    booktitle = "Proceedings of the 2022 Conference of the North American Chapter of the Association for Computational Linguistics: Human Language Technologies",
    month = jul,
    year = "2022",
    address = "Seattle, United States",
    publisher = "Association for Computational Linguistics",
    url = "https://aclanthology.org/2022.naacl-main.418",
    doi = "10.18653/v1/2022.naacl-main.418",
    pages = "5693--5709",
}

@article{tang2024tofueval,
  title={TofuEval: Evaluating Hallucinations of LLMs on Topic-Focused Dialogue Summarization},
  author={Tang, Liyan and Shalyminov, Igor and Wong, Amy Wing-mei and Burnsky, Jon and Vincent, Jake W and Yang, Yu'an and Singh, Siffi and Feng, Song and Song, Hwanjun and Su, Hang and others},
  journal={arXiv preprint arXiv:2402.13249},
  year={2024}
}

@inproceedings{tang-etal-2023-context,
    title = "In-context Learning of Large Language Models for Controlled Dialogue Summarization: A Holistic Benchmark and Empirical Analysis",
    author = "Tang, Yuting  and
      Puduppully, Ratish  and
      Liu, Zhengyuan  and
      Chen, Nancy",
    editor = "Dong, Yue  and
      Xiao, Wen  and
      Wang, Lu  and
      Liu, Fei  and
      Carenini, Giuseppe",
    booktitle = "Proceedings of the 4th New Frontiers in Summarization Workshop",
    month = dec,
    year = "2023",
    address = "Singapore",
    publisher = "Association for Computational Linguistics",
    url = "https://aclanthology.org/2023.newsum-1.6",
    doi = "10.18653/v1/2023.newsum-1.6",
    pages = "56--67",
}

@article{ramprasad2024analyzing,
  title={Analyzing LLM Behavior in Dialogue Summarization: Unveiling Circumstantial Hallucination Trends},
  author={Ramprasad, Sanjana and Ferracane, Elisa and Lipton, Zachary C},
  journal={arXiv preprint arXiv:2406.03487},
  year={2024}
}

@inproceedings{liu-etal-2023-revisiting,
    title = "Revisiting the Gold Standard: Grounding Summarization Evaluation with Robust Human Evaluation",
    author = "Liu, Yixin  and
      Fabbri, Alex  and
      Liu, Pengfei  and
      Zhao, Yilun  and
      Nan, Linyong  and
      Han, Ruilin  and
      Han, Simeng  and
      Joty, Shafiq  and
      Wu, Chien-Sheng  and
      Xiong, Caiming  and
      Radev, Dragomir",
    editor = "Rogers, Anna  and
      Boyd-Graber, Jordan  and
      Okazaki, Naoaki",
    booktitle = "Proceedings of the 61st Annual Meeting of the Association for Computational Linguistics (Volume 1: Long Papers)",
    month = jul,
    year = "2023",
    address = "Toronto, Canada",
    publisher = "Association for Computational Linguistics",
    url = "https://aclanthology.org/2023.acl-long.228/",
    doi = "10.18653/v1/2023.acl-long.228",
    pages = "4140--4170"
}

@inproceedings{min-etal-2023-factscore,
    title = "{FA}ct{S}core: Fine-grained Atomic Evaluation of Factual Precision in Long Form Text Generation",
    author = "Min, Sewon  and
      Krishna, Kalpesh  and
      Lyu, Xinxi  and
      Lewis, Mike  and
      Yih, Wen-tau  and
      Koh, Pang  and
      Iyyer, Mohit  and
      Zettlemoyer, Luke  and
      Hajishirzi, Hannaneh",
    editor = "Bouamor, Houda  and
      Pino, Juan  and
      Bali, Kalika",
    booktitle = "Proceedings of the 2023 Conference on Empirical Methods in Natural Language Processing",
    month = dec,
    year = "2023",
    address = "Singapore",
    publisher = "Association for Computational Linguistics",
    url = "https://aclanthology.org/2023.emnlp-main.741/",
    doi = "10.18653/v1/2023.emnlp-main.741",
    pages = "12076--12100"
}

@inproceedings{tang-etal-2024-minicheck,
    title = "{M}ini{C}heck: Efficient Fact-Checking of {LLM}s on Grounding Documents",
    author = "Tang, Liyan  and
      Laban, Philippe  and
      Durrett, Greg",
    editor = "Al-Onaizan, Yaser  and
      Bansal, Mohit  and
      Chen, Yun-Nung",
    booktitle = "Proceedings of the 2024 Conference on Empirical Methods in Natural Language Processing",
    month = nov,
    year = "2024",
    address = "Miami, Florida, USA",
    publisher = "Association for Computational Linguistics",
    url = "https://aclanthology.org/2024.emnlp-main.499/",
    doi = "10.18653/v1/2024.emnlp-main.499",
    pages = "8818--8847"
}

@misc{wei2024longformfactualitylargelanguage,
      title={Long-form factuality in large language models}, 
      author={Jerry Wei and Chengrun Yang and Xinying Song and Yifeng Lu and Nathan Hu and Jie Huang and Dustin Tran and Daiyi Peng and Ruibo Liu and Da Huang and Cosmo Du and Quoc V. Le},
      year={2024},
      eprint={2403.18802},
      archivePrefix={arXiv},
      primaryClass={cs.CL},
      url={https://arxiv.org/abs/2403.18802},
}

@inproceedings{song-etal-2024-veriscore,
    title = "{V}eri{S}core: Evaluating the factuality of verifiable claims in long-form text generation",
    author = "Song, Yixiao  and
      Kim, Yekyung  and
      Iyyer, Mohit",
    editor = "Al-Onaizan, Yaser  and
      Bansal, Mohit  and
      Chen, Yun-Nung",
    booktitle = "Findings of the Association for Computational Linguistics: EMNLP 2024",
    month = nov,
    year = "2024",
    address = "Miami, Florida, USA",
    publisher = "Association for Computational Linguistics",
    url = "https://aclanthology.org/2024.findings-emnlp.552/",
    doi = "10.18653/v1/2024.findings-emnlp.552",
    pages = "9447--9474"
}

@inproceedings{song-etal-2024-finesure,
    title = "{F}ine{S}ur{E}: Fine-grained Summarization Evaluation using {LLM}s",
    author = "Song, Hwanjun  and
      Su, Hang  and
      Shalyminov, Igor  and
      Cai, Jason  and
      Mansour, Saab",
    editor = "Ku, Lun-Wei  and
      Martins, Andre  and
      Srikumar, Vivek",
    booktitle = "Proceedings of the 62nd Annual Meeting of the Association for Computational Linguistics (Volume 1: Long Papers)",
    month = aug,
    year = "2024",
    address = "Bangkok, Thailand",
    publisher = "Association for Computational Linguistics",
    url = "https://aclanthology.org/2024.acl-long.51/",
    doi = "10.18653/v1/2024.acl-long.51",
    pages = "906--922"
}

@inproceedings{lee-etal-2024-unisumeval,
    title = "{U}ni{S}um{E}val: Towards Unified, Fine-grained, Multi-dimensional Summarization Evaluation for {LLM}s",
    author = "Lee, Yuho  and
      Yun, Taewon  and
      Cai, Jason  and
      Su, Hang  and
      Song, Hwanjun",
    editor = "Al-Onaizan, Yaser  and
      Bansal, Mohit  and
      Chen, Yun-Nung",
    booktitle = "Findings of the Association for Computational Linguistics: EMNLP 2024",
    month = nov,
    year = "2024",
    address = "Miami, Florida, USA",
    publisher = "Association for Computational Linguistics",
    url = "https://aclanthology.org/2024.findings-emnlp.227/",
    doi = "10.18653/v1/2024.findings-emnlp.227",
    pages = "3941--3960"
}

@misc{li2023structuredchainofthoughtpromptingcode,
      title={Structured Chain-of-Thought Prompting for Code Generation}, 
      author={Jia Li and Ge Li and Yongmin Li and Zhi Jin},
      year={2023},
      eprint={2305.06599},
      archivePrefix={arXiv},
      primaryClass={cs.SE},
      url={https://arxiv.org/abs/2305.06599}, 
}

@inproceedings{NEURIPS2023_1b44b878,
 author = {Shinn, Noah and Cassano, Federico and Gopinath, Ashwin and Narasimhan, Karthik and Yao, Shunyu},
 booktitle = {Advances in Neural Information Processing Systems},
 editor = {A. Oh and T. Naumann and A. Globerson and K. Saenko and M. Hardt and S. Levine},
 pages = {8634--8652},
 publisher = {Curran Associates, Inc.},
 title = {Reflexion: language agents with verbal reinforcement learning},
 url = {https://proceedings.neurips.cc/paper_files/paper/2023/file/1b44b878bb782e6954cd888628510e90-Paper-Conference.pdf},
 volume = {36},
 year = {2023}
}

@inproceedings{NEURIPS2022_9d560961,
 author = {Wei, Jason and Wang, Xuezhi and Schuurmans, Dale and Bosma, Maarten and ichter, brian and Xia, Fei and Chi, Ed and Le, Quoc V and Zhou, Denny},
 booktitle = {Advances in Neural Information Processing Systems},
 editor = {S. Koyejo and S. Mohamed and A. Agarwal and D. Belgrave and K. Cho and A. Oh},
 pages = {24824--24837},
 publisher = {Curran Associates, Inc.},
 title = {Chain-of-Thought Prompting Elicits Reasoning in Large Language Models},
 url = {https://proceedings.neurips.cc/paper_files/paper/2022/file/9d5609613524ecf4f15af0f7b31abca4-Paper-Conference.pdf},
 volume = {35},
 year = {2022}
}

@inproceedings{wang2024improving,
  title={Improving the robustness of knowledge-grounded dialogue via contrastive learning},
  author={Wang, Jiaan and Qu, Jianfeng and Wang, Kexin and Li, Zhixu and Hua, Wen and Li, Ximing and Liu, An},
  booktitle={Proceedings of the AAAI Conference on Artificial Intelligence},
  volume={38},
  number={17},
  pages={19135--19143},
  year={2024}
}

@inproceedings{wang-etal-2023-large,
    title = "Large Language Models as Source Planner for Personalized Knowledge-grounded Dialogues",
    author = "Wang, Hongru  and
      Hu, Minda  and
      Deng, Yang  and
      Wang, Rui  and
      Mi, Fei  and
      Wang, Weichao  and
      Wang, Yasheng  and
      Kwan, Wai-Chung  and
      King, Irwin  and
      Wong, Kam-Fai",
    editor = "Bouamor, Houda  and
      Pino, Juan  and
      Bali, Kalika",
    booktitle = "Findings of the Association for Computational Linguistics: EMNLP 2023",
    month = dec,
    year = "2023",
    address = "Singapore",
    publisher = "Association for Computational Linguistics",
    url = "https://aclanthology.org/2023.findings-emnlp.641/",
    doi = "10.18653/v1/2023.findings-emnlp.641",
    pages = "9556--9569"
}

@inproceedings{tian-etal-2024-dialogue,
    title = "Dialogue Summarization with Mixture of Experts based on Large Language Models",
    author = "Tian, Yuanhe  and
      Xia, Fei  and
      Song, Yan",
    editor = "Ku, Lun-Wei  and
      Martins, Andre  and
      Srikumar, Vivek",
    booktitle = "Proceedings of the 62nd Annual Meeting of the Association for Computational Linguistics (Volume 1: Long Papers)",
    month = aug,
    year = "2024",
    address = "Bangkok, Thailand",
    publisher = "Association for Computational Linguistics",
    url = "https://aclanthology.org/2024.acl-long.385/",
    doi = "10.18653/v1/2024.acl-long.385",
    pages = "7143--7155"
}

@inproceedings{chen-etal-2023-exploring-use,
    title = "Exploring the Use of Large Language Models for Reference-Free Text Quality Evaluation: An Empirical Study",
    author = "Chen, Yi  and
      Wang, Rui  and
      Jiang, Haiyun  and
      Shi, Shuming  and
      Xu, Ruifeng",
    editor = "Park, Jong C.  and
      Arase, Yuki  and
      Hu, Baotian  and
      Lu, Wei  and
      Wijaya, Derry  and
      Purwarianti, Ayu  and
      Krisnadhi, Adila Alfa",
    booktitle = "Findings of the Association for Computational Linguistics: IJCNLP-AACL 2023 (Findings)",
    month = nov,
    year = "2023",
    address = "Nusa Dua, Bali",
    publisher = "Association for Computational Linguistics",
    url = "https://aclanthology.org/2023.findings-ijcnlp.32/",
    doi = "10.18653/v1/2023.findings-ijcnlp.32",
    pages = "361--374"
}

@inproceedings{lin-etal-2021-csds,
    title = "{CSDS}: A Fine-Grained {C}hinese Dataset for Customer Service Dialogue Summarization",
    author = "Lin, Haitao  and
      Ma, Liqun  and
      Zhu, Junnan  and
      Xiang, Lu  and
      Zhou, Yu  and
      Zhang, Jiajun  and
      Zong, Chengqing",
    editor = "Moens, Marie-Francine  and
      Huang, Xuanjing  and
      Specia, Lucia  and
      Yih, Scott Wen-tau",
    booktitle = "Proceedings of the 2021 Conference on Empirical Methods in Natural Language Processing",
    month = nov,
    year = "2021",
    address = "Online and Punta Cana, Dominican Republic",
    publisher = "Association for Computational Linguistics",
    url = "https://aclanthology.org/2021.emnlp-main.365",
    doi = "10.18653/v1/2021.emnlp-main.365",
    pages = "4436--4451",
}

@inproceedings{zhong2022dialoglm,
  title={Dialoglm: Pre-trained model for long dialogue understanding and summarization},
  author={Zhong, Ming and Liu, Yang and Xu, Yichong and Zhu, Chenguang and Zeng, Michael},
  booktitle={Proceedings of the AAAI Conference on Artificial Intelligence},
  volume={36},
  number={10},
  pages={11765--11773},
  year={2022}
}

@inproceedings{zou-etal-2021-low,
    title = "Low-Resource Dialogue Summarization with Domain-Agnostic Multi-Source Pretraining",
    author = "Zou, Yicheng  and
      Zhu, Bolin  and
      Hu, Xingwu  and
      Gui, Tao  and
      Zhang, Qi",
    editor = "Moens, Marie-Francine  and
      Huang, Xuanjing  and
      Specia, Lucia  and
      Yih, Scott Wen-tau",
    booktitle = "Proceedings of the 2021 Conference on Empirical Methods in Natural Language Processing",
    month = nov,
    year = "2021",
    address = "Online and Punta Cana, Dominican Republic",
    publisher = "Association for Computational Linguistics",
    url = "https://aclanthology.org/2021.emnlp-main.7/",
    doi = "10.18653/v1/2021.emnlp-main.7",
    pages = "80--91"
}

@inproceedings{fang-etal-2022-spoken,
    title = "From spoken dialogue to formal summary: An utterance rewriting for dialogue summarization",
    author = "Fang, Yue  and
      Zhang, Hainan  and
      Chen, Hongshen  and
      Ding, Zhuoye  and
      Long, Bo  and
      Lan, Yanyan  and
      Zhou, Yanquan",
    editor = "Carpuat, Marine  and
      de Marneffe, Marie-Catherine  and
      Meza Ruiz, Ivan Vladimir",
    booktitle = "Proceedings of the 2022 Conference of the North American Chapter of the Association for Computational Linguistics: Human Language Technologies",
    month = jul,
    year = "2022",
    address = "Seattle, United States",
    publisher = "Association for Computational Linguistics",
    url = "https://aclanthology.org/2022.naacl-main.283/",
    doi = "10.18653/v1/2022.naacl-main.283",
    pages = "3859--3869"
}

@inproceedings{chen-yang-2021-structure,
    title = "Structure-Aware Abstractive Conversation Summarization via Discourse and Action Graphs",
    author = "Chen, Jiaao  and
      Yang, Diyi",
    editor = "Toutanova, Kristina  and
      Rumshisky, Anna  and
      Zettlemoyer, Luke  and
      Hakkani-Tur, Dilek  and
      Beltagy, Iz  and
      Bethard, Steven  and
      Cotterell, Ryan  and
      Chakraborty, Tanmoy  and
      Zhou, Yichao",
    booktitle = "Proceedings of the 2021 Conference of the North American Chapter of the Association for Computational Linguistics: Human Language Technologies",
    month = jun,
    year = "2021",
    address = "Online",
    publisher = "Association for Computational Linguistics",
    url = "https://aclanthology.org/2021.naacl-main.109/",
    doi = "10.18653/v1/2021.naacl-main.109",
    pages = "1380--1391"
}

@inproceedings{wang-etal-2022-analyzing,
    title = "Analyzing and Evaluating Faithfulness in Dialogue Summarization",
    author = "Wang, Bin  and
      Zhang, Chen  and
      Zhang, Yan  and
      Chen, Yiming  and
      Li, Haizhou",
    editor = "Goldberg, Yoav  and
      Kozareva, Zornitsa  and
      Zhang, Yue",
    booktitle = "Proceedings of the 2022 Conference on Empirical Methods in Natural Language Processing",
    month = dec,
    year = "2022",
    address = "Abu Dhabi, United Arab Emirates",
    publisher = "Association for Computational Linguistics",
    url = "https://aclanthology.org/2022.emnlp-main.325/",
    doi = "10.18653/v1/2022.emnlp-main.325",
    pages = "4897--4908"
}

@article{lu2025mutual,
  title={Mutual Reinforcement of LLM Dialogue Synthesis and Summarization Capabilities for Few-Shot Dialogue Summarization},
  author={Lu, Yen-Ju and Hu, Ting-Yao and Koppula, Hema Swetha and Pouransari, Hadi and Chang, Jen-Hao Rick and Xia, Yin and Kong, Xiang and Zhu, Qi and Wang, Simon and Tuzel, Oncel and others},
  journal={arXiv preprint arXiv:2502.17328},
  year={2025}
}

@inproceedings{zhu-etal-2023-annotating,
    title = "Annotating and Detecting Fine-grained Factual Errors for Dialogue Summarization",
    author = "Zhu, Rongxin  and
      Qi, Jianzhong  and
      Lau, Jey Han",
    editor = "Rogers, Anna  and
      Boyd-Graber, Jordan  and
      Okazaki, Naoaki",
    booktitle = "Proceedings of the 61st Annual Meeting of the Association for Computational Linguistics (Volume 1: Long Papers)",
    month = jul,
    year = "2023",
    address = "Toronto, Canada",
    publisher = "Association for Computational Linguistics",
    url = "https://aclanthology.org/2023.acl-long.377/",
    doi = "10.18653/v1/2023.acl-long.377",
    pages = "6825--6845"
}

@inproceedings{zhu-etal-2025-factual,
    title = "Factual Dialogue Summarization via Learning from Large Language Models",
    author = "Zhu, Rongxin  and
      Lau, Jey Han  and
      Qi, Jianzhong",
    editor = "Rambow, Owen  and
      Wanner, Leo  and
      Apidianaki, Marianna  and
      Al-Khalifa, Hend  and
      Eugenio, Barbara Di  and
      Schockaert, Steven",
    booktitle = "Proceedings of the 31st International Conference on Computational Linguistics",
    month = jan,
    year = "2025",
    address = "Abu Dhabi, UAE",
    publisher = "Association for Computational Linguistics",
    url = "https://aclanthology.org/2025.coling-main.302/",
    pages = "4474--4492"
}

@article{liu2024exploring,
  title={Exploring the potential of ChatGPT in medical dialogue summarization: a study on consistency with human preferences},
  author={Liu, Yong and Ju, Shenggen and Wang, Junfeng},
  journal={BMC Medical Informatics and Decision Making},
  volume={24},
  number={1},
  pages={75},
  year={2024},
  publisher={Springer}
}

@inproceedings{zhang-etal-2020-grounded,
    title = "Grounded Conversation Generation as Guided Traverses in Commonsense Knowledge Graphs",
    author = "Zhang, Houyu  and
      Liu, Zhenghao  and
      Xiong, Chenyan  and
      Liu, Zhiyuan",
    editor = "Jurafsky, Dan  and
      Chai, Joyce  and
      Schluter, Natalie  and
      Tetreault, Joel",
    booktitle = "Proceedings of the 58th Annual Meeting of the Association for Computational Linguistics",
    month = jul,
    year = "2020",
    address = "Online",
    publisher = "Association for Computational Linguistics",
    url = "https://aclanthology.org/2020.acl-main.184/",
    doi = "10.18653/v1/2020.acl-main.184",
    pages = "2031--2043"
}

@inproceedings{NEURIPS2020_6b493230,
 author = {Lewis, Patrick and Perez, Ethan and Piktus, Aleksandra and Petroni, Fabio and Karpukhin, Vladimir and Goyal, Naman and K\"{u}ttler, Heinrich and Lewis, Mike and Yih, Wen-tau and Rockt\"{a}schel, Tim and Riedel, Sebastian and Kiela, Douwe},
 booktitle = {Advances in Neural Information Processing Systems},
 editor = {H. Larochelle and M. Ranzato and R. Hadsell and M.F. Balcan and H. Lin},
 pages = {9459--9474},
 publisher = {Curran Associates, Inc.},
 title = {Retrieval-Augmented Generation for Knowledge-Intensive NLP Tasks},
 url = {https://proceedings.neurips.cc/paper_files/paper/2020/file/6b493230205f780e1bc26945df7481e5-Paper.pdf},
 volume = {33},
 year = {2020}
}

@inproceedings{chen2024benchmarking,
  title={Benchmarking large language models in retrieval-augmented generation},
  author={Chen, Jiawei and Lin, Hongyu and Han, Xianpei and Sun, Le},
  booktitle={Proceedings of the AAAI Conference on Artificial Intelligence},
  volume={38},
  number={16},
  pages={17754--17762},
  year={2024}
}

@article{kirstein2024cads,
  title={CADS: A Systematic Literature Review on the Challenges of Abstractive Dialogue Summarization},
  author={Kirstein, Frederic and Wahle, Jan Philip and Gipp, Bela and Ruas, Terry},
  journal={arXiv preprint arXiv:2406.07494},
  year={2024}
}

@article{goyal2022news,
  title={News summarization and evaluation in the era of gpt-3},
  author={Goyal, Tanya and Li, Junyi Jessy and Durrett, Greg},
  journal={arXiv preprint arXiv:2209.12356},
  year={2022}
}

@article{zhang2024benchmarking,
  title={Benchmarking large language models for news summarization},
  author={Zhang, Tianyi and Ladhak, Faisal and Durmus, Esin and Liang, Percy and McKeown, Kathleen and Hashimoto, Tatsunori B},
  journal={Transactions of the Association for Computational Linguistics},
  volume={12},
  pages={39--57},
  year={2024},
  publisher={MIT Press One Broadway, 12th Floor, Cambridge, Massachusetts 02142, USA~…}
}

@inproceedings{liu-etal-2024-benchmarking,
    title = "Benchmarking Generation and Evaluation Capabilities of Large Language Models for Instruction Controllable Summarization",
    author = "Liu, Yixin  and
      Fabbri, Alexander  and
      Chen, Jiawen  and
      Zhao, Yilun  and
      Han, Simeng  and
      Joty, Shafiq  and
      Liu, Pengfei  and
      Radev, Dragomir  and
      Wu, Chien-Sheng  and
      Cohan, Arman",
    editor = "Duh, Kevin  and
      Gomez, Helena  and
      Bethard, Steven",
    booktitle = "Findings of the Association for Computational Linguistics: NAACL 2024",
    month = jun,
    year = "2024",
    address = "Mexico City, Mexico",
    publisher = "Association for Computational Linguistics",
    url = "https://aclanthology.org/2024.findings-naacl.280/",
    doi = "10.18653/v1/2024.findings-naacl.280",
    pages = "4481--4501"
}

@article{joshi2015likert,
  title={Likert scale: Explored and explained},
  author={Joshi, Ankur and Kale, Saket and Chandel, Satish and Pal, D Kumar},
  journal={British journal of applied science \& technology},
  volume={7},
  number={4},
  pages={396},
  year={2015},
  publisher={Sciencedomain International}
}

\vspace{10pt}
\appendix
\section{Paradigm-Level Performance Visualization}
\label{sec:overall_performance_visualization}
\begin{figure}[t]
  \centering
  \setlength{\abovecaptionskip}{6pt}
  \includegraphics[width=\linewidth]{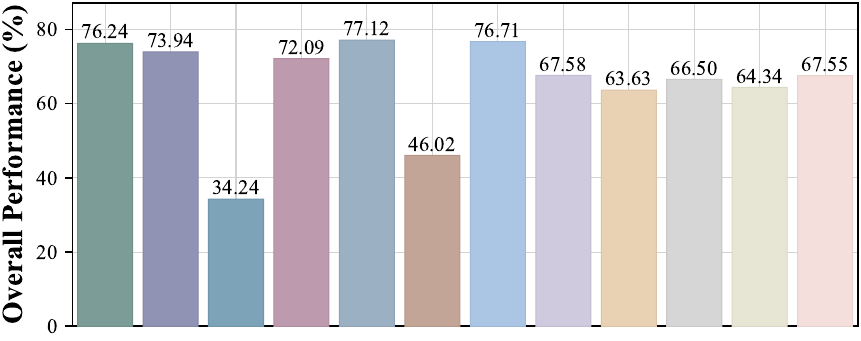}  
  \caption{Overall performance under the EBS-AOS paradigm. Models are stratified into three tiers: \textsc{Tier-1} ($[72, 77]$), \textsc{Tier-2} ($[64, 68]$), and \textsc{Tier-3} ($[34, 46]$).}
  \label{fig:op_ebs-aos}
  \vspace{-2pt}
\end{figure}

\begin{figure}[t]
  \centering
  \setlength{\abovecaptionskip}{6pt}
  \includegraphics[width=\linewidth]{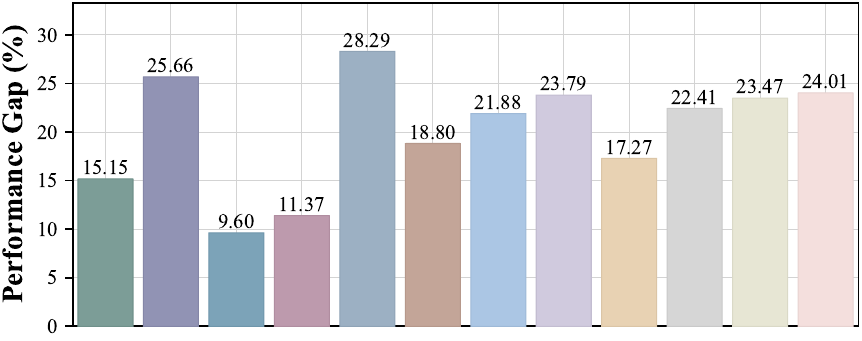}
  \caption{Overall performance gap between the two paradigms. A larger gap indicates lower stability.}
  \label{fig:op_gap-patterns}
  \vspace{-2pt}
\end{figure}

\begin{figure}[t]
  \centering
  \setlength{\abovecaptionskip}{6pt}
  \includegraphics[width=\linewidth]{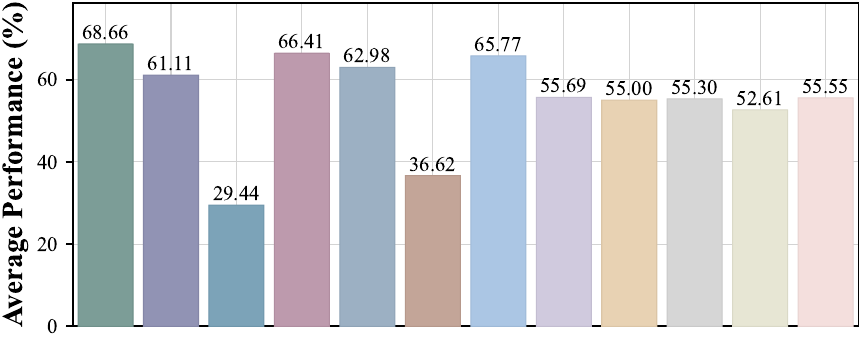}
  \caption{Average overall performance across both paradigms. Models are stratified into three tiers: \textsc{Tier-1} ($[61, 69]$), \textsc{Tier-2} ($[53, 56]$), and \textsc{Tier-3} ($[29, 37]$).}
  \label{fig:op_avg-patterns}
  \vspace{-2pt}
\end{figure}

Figure~\ref{fig:op_ebs-aos} and Figure~\ref{fig:op_abs-aos} illustrate the overall performance of each model under the EBS-AOS and ABS-AOS paradigms, respectively. Figure~\ref{fig:op_avg-patterns} presents the average performance of each model across both paradigms. Figure~\ref{fig:op_gap-patterns} reveals the cross-paradigm stability of each model by quantifying the performance gap between the two paradigms.

\begin{figure*}[t]
  \centering
  \setlength{\abovecaptionskip}{0pt}
  \includegraphics[width=0.916\linewidth]{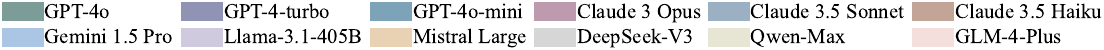}
  \label{fig:model_name}
  \vspace{-10pt}
\end{figure*}

\section{Further Analysis}
\label{sec:further_analysis}
\paragraph{LLMs are better retrievers than generators for background summary.} As presented in Table~\ref{tab:main_evaluation_results}, all LLMs exhibit higher $\mathrm{SP}$ metrics compared to their $\mathrm{KSAF}$ counterparts. This gap between retrieval and generation indicates that LLMs possess stronger in-context recognition capabilities for coarse-grained paragraphs than fine-grained facts.

\paragraph{Cross-paradigm stability of LLMs.} By analyzing the performance gap between the two summarization paradigms (see Figures~\ref{fig:op_ebs-aos}, \ref{fig:op_gap-patterns}, and \ref{fig:op_abs-aos} in Appendix~\ref{sec:overall_performance_visualization}), we find that different LLMs excel in different paradigms. For example, Claude 3.5 Sonnet exhibits a significant gap ($28.29$\scalebox{0.90}{$\%$}), while Claude 3 Opus demonstrates a relatively smaller gap ($11.37$\scalebox{0.90}{$\%$}), indicating stronger stability.

\paragraph{LLMs perform better with EBS-AOS than ABS-AOS for KGDS.} \enspace As presented in Table~\ref{tab:main_evaluation_results}, all LLMs achieve superior $\mathrm{PLQ}_{\mathrm{GM}}$ in the EBS-AOS paradigm compared to ABS-AOS, due to the more complete and accurate background summaries and higher-quality opinion summaries. Therefore, we suggest prioritizing EBS-AOS in real-world KGDS for more effective implementation.

\paragraph{Self-reflection makes LLMs more conservative or open for background summary.} \enspace As shown in Table~\ref{tab:main_evaluation_results}, most LLMs demonstrate polarization in the increases and decreases between $\mathrm{SP}_{\mathrm{R}}$ and $\mathrm{SP}_{\mathrm{P}}$, as well as $\mathrm{KSAF}_{\mathrm{R}}$ and $\mathrm{KSAF}_{\mathrm{P}}$. This suggests different reflection strategies: some prioritize precision (e.g., DeepSeek-V3 with $\mathrm{SP}_{\mathrm{R}}$ \textcolor{red}{${\downarrow 1.19}$}, $\mathrm{SP}_{\mathrm{P}}$ \textcolor{green}{${\uparrow 1.37}$}), while others prioritize recall (e.g., Claude 3 Opus). A few models exhibit simultaneous increases (e.g., GPT-4o with $\mathrm{SP}_{\mathrm{R}}$ \textcolor{green}{${\uparrow 1.56}$}, $\mathrm{SP}_{\mathrm{P}}$ \textcolor{green}{${\uparrow 0.34}$}) or decreases (e.g., Qwen-Max), indicating more balanced or weaker reflection abilities.

\begin{figure}[t]
  \centering
  \setlength{\abovecaptionskip}{6pt}
  \includegraphics[width=\linewidth]{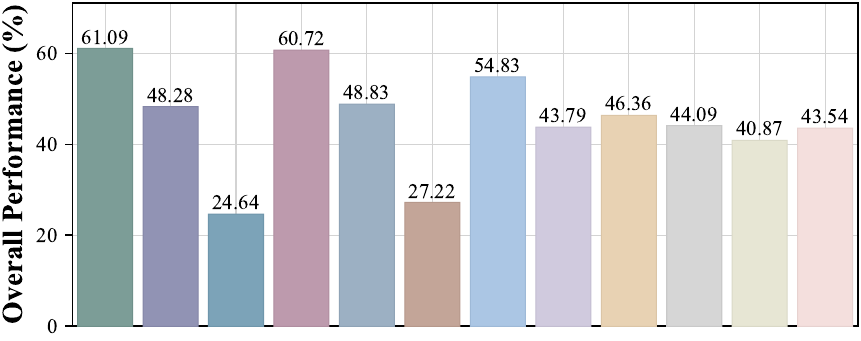}  
  \caption{Overall performance under the ABS-AOS paradigm. Models are stratified into three tiers: \textsc{Tier-1} ($[55, 61]$), \textsc{Tier-2} ($[41, 49]$), and \textsc{Tier-3} ($[25, 27]$).}
  \label{fig:op_abs-aos}
  \vspace{-2pt}
\end{figure}

\section{Detailed Formulas of Evaluation Metrics}
\subsection{ABS Evaluation Metrics}
\label{sec:abs_evaluation_metrics}
Let $\mathcal{B}$ denote the LLM-generated ABS, and let $\mathcal{K}=\{k_i\}_{i=1}^{m}$ and $\mathcal{N}=\{n_j\}_{j=1}^{p}$ represent the sets of $m$ key supporting and $p$ nonsupporting atomic facts, respectively. We define the fact verification function $\phi$ as:
\begin{equation}
\phi:(\mathcal{B}, f) \rightarrow \{0, 1\}, \; f \in \mathcal{K} \cup \mathcal{N}
\end{equation}
where $\phi(\mathcal{B}, f) = 1$ if the fact $f$ can be inferred from $\mathcal{B}$, and $0$ otherwise.

\paragraph{KSAF Recall.} This metric quantifies the \textbf{coverage} of key supporting atomic facts in the ABS:
\begin{equation}
\mathrm{KSAF}_{\mathrm{R}} = \frac{1}{m} \sum_{i=1}^{m} \phi(\mathcal{B}, k_i)
\end{equation}

\paragraph{KSAF Precision.} This metric measures the \textbf{focus} of the ABS on key supporting atomic facts:
\begin{equation}
\mathrm{KSAF}_{\mathrm{P}} = \frac{\sum_{k \in \mathcal{K}} \phi(\mathcal{B}, k)}{\sum_{f \in \mathcal{K} \cup \mathcal{N}} \phi(\mathcal{B}, f)}
\end{equation}

\paragraph{KSAF F1-score.} This metric evaluates the \textbf{overall quality} of the ABS by comprehensively considering both coverage and focus:
\begin{equation}
\mathrm{KSAF}_{\mathrm{F_1}} = 2 \cdot \frac{\mathrm{Precision} \cdot \mathrm{Recall}}{\mathrm{Precision} + \mathrm{Recall}}
\end{equation}

\subsection{AOS Evaluation Metrics}
\label{sec:aos_evaluation_metrics}
Let $\mathcal{O}$ denote the LLM-generated AOS, and let $\mathcal{C}=\{c_i\}_{i=1}^{n}$ represent the set of $n$ clear atomic opinions. We define the opinion verification function $\psi$ as:
\begin{equation}
\psi:(\mathcal{O}, c) \rightarrow \{0, 1\}, \; c \in \mathcal{C}
\end{equation}
where $\psi(\mathcal{O}, c) = 1$ if the opinion $c$ can be inferred from $\mathcal{O}$, and $0$ otherwise.

\paragraph{CAO Recall.} This metric quantifies the \textbf{coverage} of clear atomic opinions in the AOS:
\begin{equation}
\mathrm{CAO}_{\mathrm{R}} = \frac{1}{n} \sum_{i=1}^{n} \psi(\mathcal{O}, c_i)
\end{equation}

\section{Detailed Benchmark Statistics}
\label{sec:statistics_for_kgds_benchmark}

\paragraph{News Statistics.} Refer to Table~\ref{tab:statistics-for-sbk}.

\begin{table}[htbp]
    \centering
    \small
    \setlength{\abovecaptionskip}{10pt}
    \setlength{\tabcolsep}{4pt}
    \begin{tabular}{ccc}
        \toprule[1pt]
        \textbf{Num.} & \textbf{Paras}$_{\text{avg}}$ & \textbf{Tokens}$_{\text{avg}}$ \\
        \midrule[0.5pt]
        $100$ & $14.4$ & $617.5$ \\
        \bottomrule[1pt]
    \end{tabular}
    \caption{Statistics for news. Num. indicates the number of news articles. Paras$_{\text{avg}}$ and Tokens$_{\text{avg}}$ represent the average number of paragraphs and tokens, respectively.}
    \label{tab:statistics-for-sbk}
    \vspace{-15pt}
\end{table}

\paragraph{Discussion Statistics.} Refer to Table~\ref{tab:statistics-for-kgd}.

\begin{table}[H]
    \centering
    \small
    \setlength{\abovecaptionskip}{10pt}
    \setlength{\tabcolsep}{4pt}
    \begin{tabular}{cccc}
        \toprule[1pt]
        \textbf{Num.} & \textbf{Pts.}$_{\text{avg}}$ & \textbf{Uttrs.}$_{\text{avg}}$ & \textbf{Tokens}$_{\text{avg}}$ \\
        \midrule[0.5pt]
        $100$ & $2.0$ & $4.1$ & $112.0$ \\
        \bottomrule[1pt]
    \end{tabular}
    \caption{Statistics for discussion. Num. indicates the number of discussions. Pts.$_{\text{avg}}$, Uttrs.$_{\text{avg}}$, and Tokens$_{\text{avg}}$ represent the average number of participants, utterances, and tokens, respectively.}
    \label{tab:statistics-for-kgd}
    \vspace{-8pt}
\end{table}

\paragraph{Annotation Statistics for EBS.} Among the $1696$ paragraphs from $100$ original articles, the expert annotation consistency rate is $84.7$\scalebox{0.90}{$\%$}. A total of $1437$ paragraphs are retained for the final news articles, while $208$ are identified as ambiguous paragraphs and removed. Of the $1437$ retained, $432$ are annotated as supporting and $1005$ as nonsupporting.

\paragraph{Annotation Statistics for ABS.} Among the $2428$ atomic fact units decomposed from the $432$ supporting paragraphs, $1638$ are consistently annotated as key supporting atomic facts, $780$ as non-key atomic facts, and $10$ as intra-repetitive atomic facts.

Among the $5187$ atomic fact units from the $1005$ nonsupporting paragraphs, $4996$ are automatically annotated as nonsupporting atomic facts, $176$ as masked conflicting facts, and $15$ as intra-repetitive atomic facts.

\paragraph{Annotation Statistics for AOS.} A total of $873$ clear atomic opinions are expert manually annotated. Of these, $800$ contain clarified implicit references, while $73$ do not require clarification. Within the $800$ opinions, there are a total of $1113$ written clarified implicit references.

\section{LLM Sources}
\label{sec:llm_sources}
\textbf{OpenAI}\footnote{\url{https://platform.openai.com/docs/api-reference/introduction}}: GPT-4o, GPT-4-turbo, GPT-4o-mini

\vspace{5pt}

\noindent\textbf{Anthropic}\footnote{\url{https://docs.anthropic.com/en/api/getting-started}}: Claude 3 Opus, Claude 3.5 Sonnet, Claude 3.5 Haiku

\vspace{5pt}

\noindent\textbf{Google}\footnote{\url{https://ai.google.dev/gemini-api/docs}}: Gemini 1.5 Pro

\vspace{5pt}

\noindent\textbf{Meta}\footnote{\url{https://www.llmapi.com}}: Llama-3.1-405B

\vspace{5pt}

\noindent\textbf{Mistral AI}\footnote{\url{https://docs.mistral.ai/api/}}: Mistral Large

\vspace{5pt}

\noindent\textbf{DeepSeek}\footnote{\url{https://api-docs.deepseek.com/zh-cn/}}: DeepSeek-V3

\vspace{5pt}

\noindent\textbf{Alibaba}\footnote{\url{https://bailian.console.aliyun.com/}}: Qwen-Max

\vspace{5pt}

\noindent\textbf{ZhipuAI}\footnote{\url{https://www.bigmodel.cn/dev/api/normal-model/glm-4}}: GLM-4-Plus

\section{Human Evaluation Instruction}
\begin{quote}
\textit{You will be given a pair of summaries (a Background Summary and an Opinion Summary).}

\textit{Please evaluate the overall quality of this pair of summaries based on the following three criteria. Provide a single, holistic score from 1 to 5 using the Likert scale (5 = Best, 1 = Worst).}

\textit{1. Background Summary: Does the summary provide the necessary context to understand the discussion?}

\textit{2. Opinion Summary: Does the summary clearly provide the participants' viewpoints with implicit references successfully clarified?}

\textit{3. Overall Performance: Do the two summaries work together effectively for the readers?}
\end{quote}

Our human evaluation is conducted by the same PhD candidates who constructed the benchmark. We choose these experts as they possess a deep and consistent understanding of the task's goals, which is already demonstrated by their high agreement during the benchmark creation phase. This evaluation process yields a Cohen's Kappa coefficient of 0.74, indicating substantial inter-annotator agreement.

\section{Prompts and Instructions}
\label{sec:prompts_and_instructions}

\subsection{Unified Parameter Settings}
In this work, all 12 evaluated LLMs use consistent parameter settings for all prompts and instructions: \texttt{max\_token}=4096 and \texttt{temperature}=0. No other default parameters are modified.

\subsection{Atomic Fact Decomposition Instruction}
\label{sec:atomic_fact_decomposition_instruction}
\begin{figure*}[t]
  \centering
  \setlength{\abovecaptionskip}{10pt}
  \includegraphics[width=\linewidth]{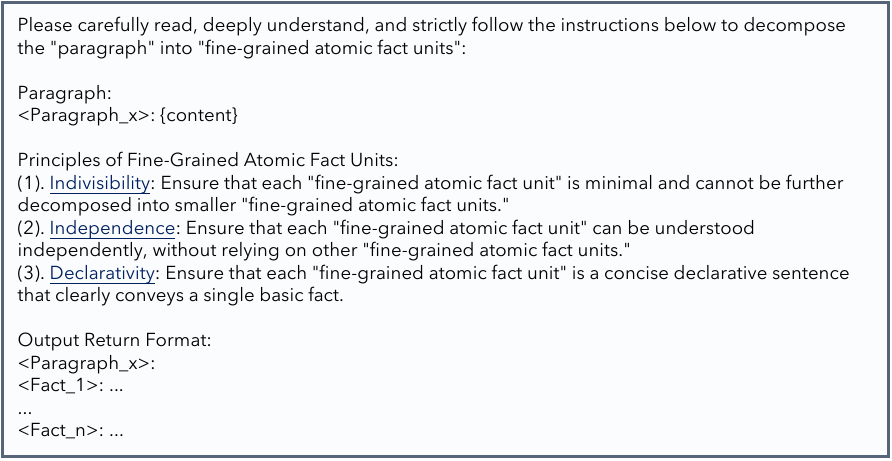}
  \caption{Atomic fact decomposition instruction.}
  \label{fig:atomic-fact-decomposition-instruction}
\end{figure*}

This instruction is used during the benchmark construction phase to guide a large language model in decomposing news paragraphs into fine-grained, atomic fact units. The instruction specifies three core principles for decomposition: indivisibility, independence, and declarativity. This ensures that the resulting fact units are minimal in granularity and informationally complete (see Figure~\ref{fig:atomic-fact-decomposition-instruction}).

\subsection{Structured Prompts}
\label{sec:structured_prompts}
\begin{figure*}[t]
  \centering
  \setlength{\abovecaptionskip}{10pt}
  \includegraphics[width=\linewidth]{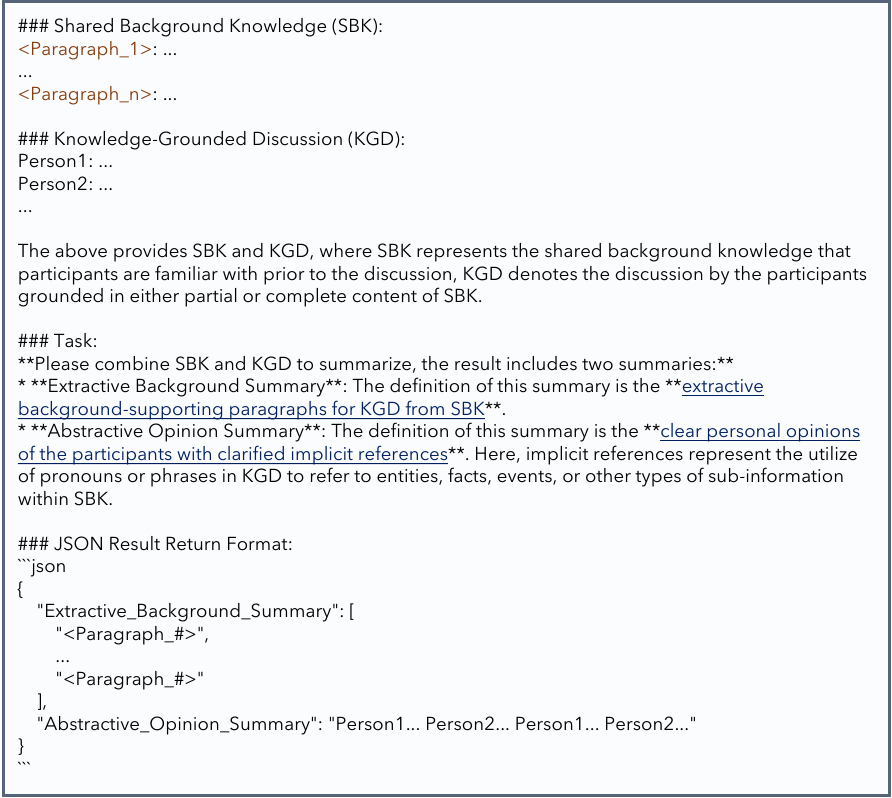}
  \caption{Structured prompt for KGDS EBS-AOS paradigm.}
  \label{fig:ebs-aos_structured-prompt}
\end{figure*}

\begin{figure*}[t]
  \centering
  \setlength{\abovecaptionskip}{10pt}
  \includegraphics[width=\linewidth]{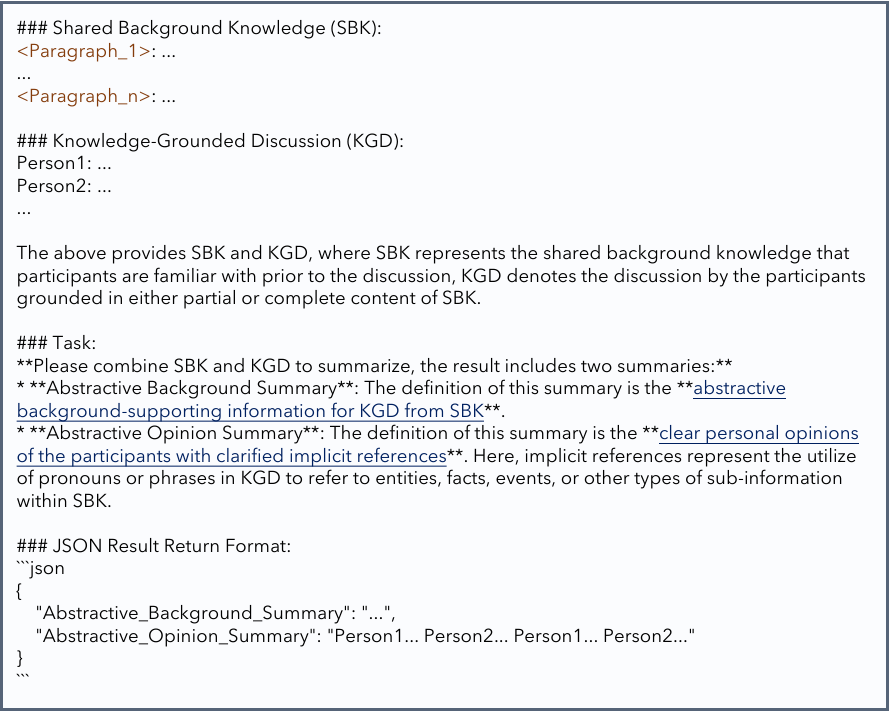}
  \caption{Structured prompt for KGDS ABS-AOS paradigm.}
  \label{fig:abs-aos_structured-prompt}
\end{figure*}

These prompts serves as the core prompt for evaluating the performance of different LLMs on the KGDS task. It provides the models with standard inputs, including the shared background knowledge and the knowledge-grounded discussion, and clearly defines the objectives and requirements for the two sub-summaries under both summarization paradigms. See Figures~\ref{fig:ebs-aos_structured-prompt} and \ref{fig:abs-aos_structured-prompt} for EBS-AOS and ABS-AOS paradigms, respectively.

\subsection{Self-Reflection Instructions}
\label{sec:self-reflection_instructions}
\begin{figure*}[t]
  \centering
  \setlength{\abovecaptionskip}{10pt}
  \includegraphics[width=\linewidth]{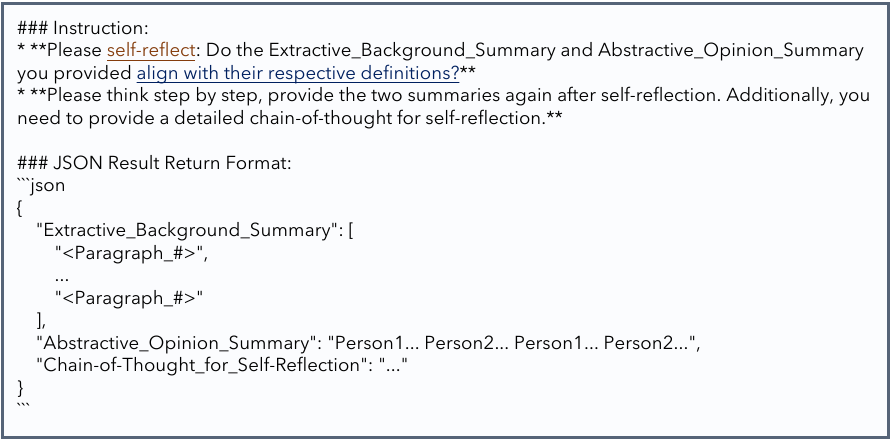}
  \caption{Self-reflection instruction for KGDS EBS-AOS paradigm.}
  \label{fig:ebs-aos_self-reflection-instruction}
\end{figure*}

\begin{figure*}[t]
  \centering
  \setlength{\abovecaptionskip}{10pt}
  \includegraphics[width=\linewidth]{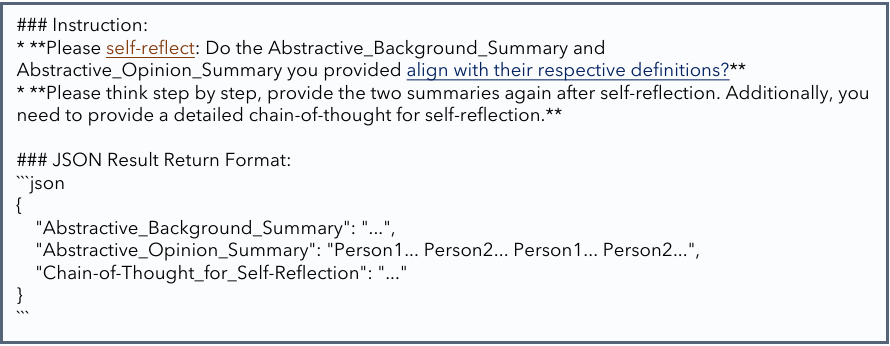}
  \caption{Self-reflection instruction for KGDS ABS-AOS paradigm.}
  \label{fig:abs-aos_self-reflection-instruction}
\end{figure*}

These instructions are used in the multi-turn self-reflection evaluation setting. After the model generates its initial summary, these instructions guide it to critically review its output, checking whether the generated background and opinion summaries strictly adhere to the task definitions. The instructions require the model to provide a detailed chain of thought in a step-by-step manner and, based on this reflection, generate a refined summary. This process is designed to explore and evaluate the model's self-correction capabilities on the KGDS task. See Figures~\ref{fig:ebs-aos_self-reflection-instruction} and \ref{fig:abs-aos_self-reflection-instruction} for the EBS-AOS and ABS-AOS paradigms, respectively.

\subsection{Verification Prompts}
\label{sec:verification_prompts}
\begin{figure*}[t]
  \centering
  \setlength{\abovecaptionskip}{10pt}
  \includegraphics[width=\linewidth]{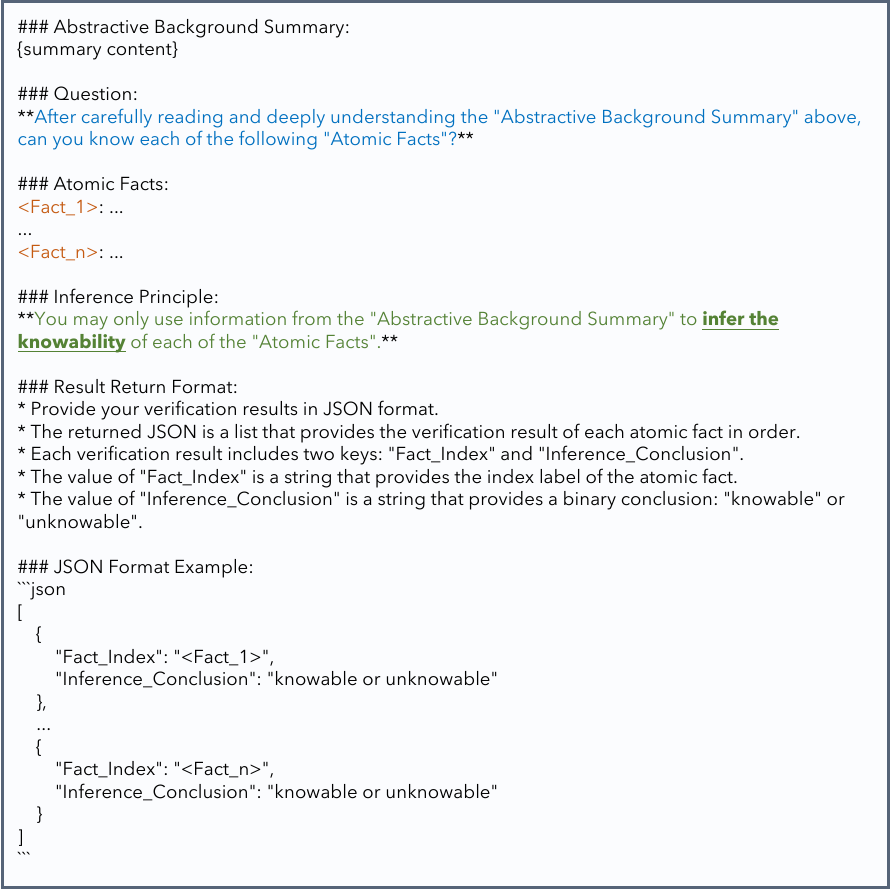}
  \caption{Atomic fact verification prompt. The number of facts ($n$) dynamically changes at the paragraph-level.}
  \label{fig:fact-verification-prompt}
\end{figure*}

\begin{figure*}[t]
  \centering
  \setlength{\abovecaptionskip}{10pt}
  \includegraphics[width=0.75\linewidth]{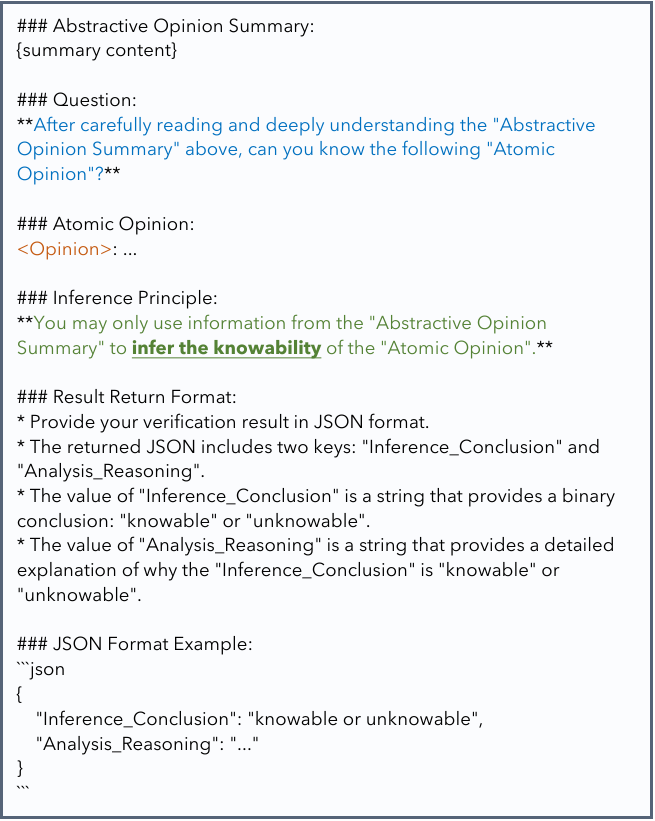}
  \caption{Atomic opinion verification prompt.}
  \label{fig:opinion-verification-prompt}
\end{figure*}

These prompts are a core component of our evaluation framework, used to automatically verify the quality of the generated summaries. We utilize a powerful verifier LLM (GPT-4o in our work) to execute these prompts, determining whether the model-generated summary contains the key information from the gold standard. Specifically, the fact verification prompt (see Figure~\ref{fig:fact-verification-prompt}) is used to assess the recall and precision of the abstractive background summary against key supporting atomic facts, while the opinion verification prompt (see Figure~\ref{fig:opinion-verification-prompt}) evaluates the coverage of the abstractive opinion summary against clear atomic opinions.

\subsection{Error Detection Instruction}
\label{sec:error_detection_instruction}
\begin{figure*}[t]
  \centering
  \setlength{\abovecaptionskip}{10pt}
  \includegraphics[width=\linewidth]{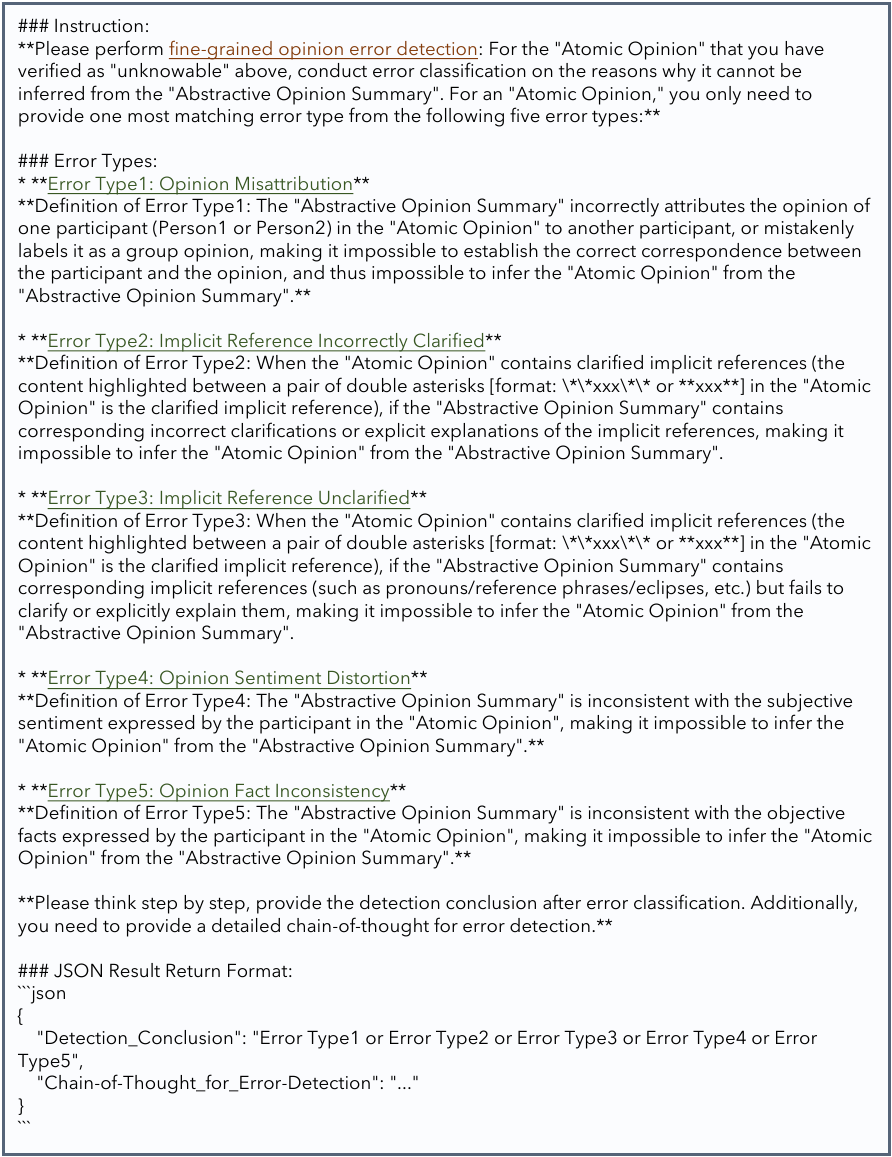}
  \caption{AOS error detection instruction.}
  \label{fig:error-detection-instruction}
\end{figure*}

This instruction is used for the fine-grained error analysis of the abstractive opinion summary. When the verifier determines that the summary has failed to cover a gold-standard clear atomic opinion, this instruction is invoked to attribute the failure to one of five predefined, specific error types (e.g., opinion misattribution, implicit reference incorrectly clarified). This approach allows us to gain deep insights into the specific weaknesses of each model in integrating opinions and clarifying references. We provide the instruction in Figure~\ref{fig:error-detection-instruction}.

\end{document}